\pdfoutput=1
\documentclass[11pt]{article}

\usepackage[preprint]{acl}

\usepackage{times}
\usepackage{titling}
\usepackage{authblk}
\usepackage{hyperref}
\usepackage{latexsym}
\usepackage{amsmath, amsthm}
\usepackage{multirow}
\usepackage{array}
\usepackage{graphicx}
\usepackage{geometry}
\usepackage[T1]{fontenc}
\usepackage[utf8]{inputenc}
\usepackage{microtype}
\usepackage{inconsolata}
\usepackage{graphicx}

\pretitle{%
    \vspace{-1cm}
    \begin{center}
    \Huge\bfseries
}
\posttitle{%
    \end{center}
    \vspace{-1cm}
}
\preauthor{%
    \begin{center}
}
\postauthor{%
    \end{center}
    \vspace{-1.5cm}
}

\title{\LARGE AgreeMate: Teaching LLMs to Haggle}
\author{
    \begin{tabular}{c@{\hskip 1cm}c@{\hskip 1cm}c}
        \textbf{Ainesh Chatterjee} & \textbf{Samuel Miller} & \textbf{Nithin Parepally} \\\vspace{-0.05em}
        \normalsize\href{mailto:aineshc@terpmail.umd.edu}{aineshc@terpmail.umd.edu} &
        \normalsize\href{mailto:samm@umd.edu}{samm@umd.edu} &
        \normalsize\href{mailto:nparepa@terpmail.umd.edu}{nparepa@terpmail.umd.edu}
    \end{tabular} \\\\
    \normalsize Department of Computer Science, University of Maryland, College Park, USA
}

\begin{document}
\maketitle


\begin{abstract}
We introduce \textbf{AgreeMate}, a framework for training Large Language Models (LLMs) to perform strategic price negotiations through natural language. We apply recent advances to a negotiation setting where two agents (i.e. buyer or seller) use natural language to bargain on goods using coarse actions. Specifically, we present the performance of Large Language Models when used as agents within a decoupled (modular) bargaining architecture. We demonstrate that using prompt engineering, fine-tuning, and chain-of-thought prompting enhances model performance, as defined by novel metrics. We use attention probing to show model attention to semantic relationships between tokens during negotiations.
\end{abstract}


\section{Introduction}
The emergence of Large Language Models (LLMs) has transformed our understanding of machine intelligence, demonstrating remarkable capabilities in language comprehension and complex social interactions. While recent works have explored LLMs' abilities in dialogue and reasoning tasks, their potential in strategic communication scenarios remains underexplored. We present AgreeMate, a comprehensive testing framework and negotiation system that explores the capabilities of LLMs in strategic price bargaining. Our approach combines role-specialized fine-tuning with systematic model comparison across scales and architectures, offering insights into how different LLM variants perform in complex negotiation scenarios.

Strategic negotiation represents a particularly challenging domain for artificial intelligence, requiring agents to simultaneously master natural language generation, maintain consistent goal-directed behavior, and adapt tactics based on counterpart responses. Previous approaches to automated negotiation often relied on complex architectures with separate modules for strategy planning and language generation \citep{he-etal-2018}. However, the emergence of instruction-tuned LLMs offers an opportunity to revisit this problem with potentially simpler, more elegant solutions that leverage these models' inherent reasoning and natural dialogue capabilities.

Our work makes several key contributions:
\begin{enumerate}
    \item We introduce a systematic framework for evaluating LLM negotiation capabilities across different model scales, architectures, and training approaches.
    \item We develop and analyze role-specialized negotiation agents (buyer, seller, generalist) through targeted fine-tuning, demonstrating the effectiveness of role-specific optimization.
    \item We present a detailed comparative analysis between base LLaMA models (3B to 70B parameters), revealing key insights about the model scale and training impacts on negotiation performance \citep{touvron2023llamaopenefficientfoundation}.
    \item We provide comprehensive evaluation metrics for negotiation success, including derived and novel measurements such as fairness, bias, relative efficiency and probing ratio.
    \item We conduct attention head probing to gain deeper insights into the internal workings of LLMs during strategic negotiation, shedding light on how specific attention mechanisms contribute to bargaining behavior.
\end{enumerate}

This project was conducted as part of the \textbf{CMSC723: Graduate Natural Language Processing (Fall 2024)} course at the University of Maryland, College Park. The content and findings reflect our independent research and do not imply any endorsement or ownership by the university.


\section{Motivation}

The capabilities of modern large language models (LLMs) have advanced significantly, with models like GPT-4 demonstrating persuasive abilities that reportedly exceed internal safety benchmarks \citep{seetharaman2024turning}. These models can craft highly effective arguments using psycholinguistic principles \citep{breum2023persuasivepowerlargelanguage}, raising important questions about their application in structured domains such as price negotiation.

Negotiation serves as an ideal testbed for LLMs, requiring a delicate balance between maximizing utility and fostering cooperation. Effective negotiation relies on implicit strategies—revealing one’s true intentions or limits typically weakens their position. This inherent opacity aligns well with the ``black box'' reasoning of LLMs, which can emulate human-like strategic decision-making without needing explicit disclosures.

Previous automated negotiation systems \citep{he-etal-2018} relied on modular architectures to separate strategic planning from natural language generation. While effective for their time, these approaches reflected the limitations of pre-transformer models, which struggled with multi-step reasoning. In contrast, instruction-tuned LLMs like LLaMA \citep{touvron2023llamaopenefficientfoundation} offer the potential to unify these tasks, discovering nuanced negotiation strategies through exposure to domain-specific dynamics rather than manual engineering.

Developing effective automated negotiation agents has practical implications. Such systems could assist in online marketplaces, enabling efficient price discovery and ensuring fair transactions by either providing strategic advice or directly participating in negotiations. These agents can address the growing need for scalable and impartial negotiation mechanisms in increasingly digital economies.

This work examines how LLM scale, post-training procedures, and role-specific fine-tuning impact negotiation performance. By comparing models ranging from 3B to 70B parameters, we aim to identify the minimal model capabilities required for effective negotiation. This research also explores how these systems can be deployed in resource-constrained settings without compromising performance, offering insights into the trade-offs between model complexity and strategic efficacy.


\section{Architecture}
Our work takes inspiration from the "Decoupling Strategy and Generation in Negotiation Dialogues" \citep{he-etal-2018} paper in which buyer and seller dialogue agents produce a distribution over a responding utterance given a dialogue history (a sequence of utterances). 

The modular framework determining an agent’s responding utterance is characterized as follows. A parser module maps input utterances into a coarse dialogue act given a dialogue history and previous dialogue acts. Acts are selected from following: greet, disagree, agree, insist, inquire, inform, counter, intro, vague-price, intro, and propose. Here, course dialogue acts do not capture the entire semantics of a sentence, which enables a rule-based approach to parsing. The parser works by detecting prices and objects referred to as “entities” and matches keyword patterns. 

Then, a manager module predicts a dialogue act given information about the scenario and past dialogue acts. The paper uses three approaches to train the manager which include supervised learning, where a sequence to sequence model with attention is used to learn transition probabilities of dialogue acts, reinforcement learning where training consists of optimizing for a reward function, and finally a hybrid approach where a learned manager is combined with hand-coded rules. For the reinforcement learning approach, three different reward functions are used. These are utility, fairness, and length. 

Finally, a generator module uses retrieval based generation, where a database of candidate utterances is searched for the best match based on the act predicted by the manager and the history, to output a responding utterance. Utterances in the database are in a template-based format where placeholders are replaced with contextually relevant information during test time. 

In our project, we adopt a similar framework to make the task suitable for an LLM. In particular, our framework eliminates the parser component and the LLM serves as both the manager and the generator. Given only a history of utterances (so no dialogue acts), scenario information, and the number of turns left in the negotiation (we enforce a limit), a buyer or seller agent selects an act from \{ intro, init-price, offer, counter-price, insist, agree, disagree, accept, inform, inquire, and unknown \}. Then, based on the selected action, a contextually appropriate utterance aligning with the action is produced. These outputs are formatted into what we call \textit{action-utterance} pairs. 

Our “manager” component does not vary with approaches like the paper which uses a supervised learning approach to mimic average human behavior and a reinforcement learning approach to directly optimize for a particular goal. Our framework provides the same set of instructions to each agent comprised of general goals like accepting reasonable offers that are close to their target prices.  

There are two reasons for crafting our framework this way. First, we wanted to observe whether an LLM can perform the work of a dedicated parser. By only providing a history of utterances, we aim to explore if LLMs can extract “entities” which may or may directly correspond to prices and objects, identify appropriate dialogue acts, and interpret the overall state of the negotiation. To goal being, to see if LLMs can infer common negotiation strategies like foot in the door or lowball from the nuances of negotiation dialogue. 

The second reason is to understand if LLMs have the capability to strategize as opposed to simply mimicking human responses. For instance, when it assumes the task of manager, is the LLM developing a high-level strategy (like countering in the current time step or dragging the conversation out through inquiry to create a better opportunity for countering) or just selecting the "next best" human-like response.


\section{Datasets}
We employed the Craiglist Negotiation Dataset \citep{he-etal-2018} and the Deal or No Deal dialogue dataset \citep{lewis2017dealdealendtoendlearning}. The Craiglist Negotiation dataset is a collection of more than 6,000 human-human dialogues where buyers and sellers negotiate on the prices of items posted on Craiglist. Postings range from popular categories such as cars, bikes and phones and each posting provides three scenarios corresponding to buyer target prices set at 50\%, 70\%, and 90\% off the listing price. Each dialogue is accompanied by a role, utterance and intent, where intent resembles a course dialogue act like propose, counter, or agree. We use this dataset to the evaluate the performance of different large language models. 

The Deal or No Deal dataset contains over 12,000 negotiation dialogues on a multi-issue bargaining task where agents who cannot observe each other’s reward functions must reach a deal via natural language dialogue. In this dataset agents negotiate on how to divide up a set of items as opposed to settling on a particular price for an item. We use this dataset for fine-tuning models into specialized roles.


\section{Fine-tuning}
We finetuned a medium-scale LLaMA-3.2-3B-Instruct language model \citep{touvron2023llamaopenefficientfoundation} into role-specific (`buyer`, `seller`) and `generalist` negotiation agents. The overarching objective was to refine role-specific and generalist LLMs to excel at price negotiation tasks, navigating strict resource constraints and complex training dynamics. Our fine-tuning approach integrates parameter-efficient techniques to achieve stable performance with limited hardware resources.

\subsection{Dataset Preparation}
We used the Deal or No Deal dataset for fine-tuning. The addition of strategic reasoning annotations (\textbf{thought process}) allowed the model to improve beyond simple mimicry. We processed the data into role-specific training splits:
\begin{itemize}
    \item \textbf{Buyer Training Set:} Focuses on scenarios where the model assumes the role of a buyer.
    \item \textbf{Seller Training Set:} Focuses on scenarios where the model assumes the role of a seller.
    \item \textbf{Generalist Training Set:} Combines both buyer and seller data to create a single model capable of handling both roles.
\end{itemize}
Each training instance consists of:
\begin{itemize}
    \item \textbf{Context:} Scenario description, listing price, and target price information.
    \item \textbf{Thought Process:} Annotations capturing strategic reasoning to improve model behavior.
    \item \textbf{Dialogue Turns:} Structured conversation history as input for fine-tuning.
\end{itemize}

\subsection{Challenges}
In fine tuning the model, we employ several techniques addressing the following challenges.
\begin{itemize}
    \item Fully fine-tuning a 3B-parameter model is computationally and memory-intensive, making it impractical on standard hardware setups. We ran into some issues working with the Nexus cluster and therefore opted to use Google Colab Plus which provided better GPUs than our local setups but it was still very limiting.

    \item Limited GPU memory necessitates memory-efficient training methodologies to handle large models and extensive datasets without encountering out-of-memory (OOM) errors.

    \item Negotiation dialogues are inherently complex, involving strategic reasoning and dynamic price adjustments. Training models on such data requires stable and sophisticated training heuristics to capture both strategy and language nuances effectively.

    \item Ensuring that the model can learn and execute negotiation strategies without entangling them with language generation is difficult. When we started this project, we attempted to finetune a smaller 1 billion parameter model. However no matter what we tried, the resulting model would constantly inject the prompt and its strategy into utterances, making it infeasible to build reasonable dialogue histories.
\end{itemize}

\subsection{Training \& Techniques}
We first implemented Low-Rank Adaptation (LoRA) to fine-tune the model. LoRA reduces the number of trainable parameters by introducing low-rank matrices into attention and feedforward layers. This method allows the model to adapt to negotiation tasks without updating all original parameters, significantly reducing memory usage and computational overhead.  We applied LoRA to the \texttt{q\_proj}, \texttt{k\_proj}, \texttt{v\_proj}, and \texttt{o\_proj} modules with a rank of 32 and an alpha of 16.

Then, we applied 4-bit quantization with the nf4 quant type to compress the model weights. This drastic reduction in memory footprint enables the training of large models on hardware with limited GPU memory, while maintaining performance levels comparable to higher-precision models.
Memory usage stabilized at approximately 3.56 GB, enabling training on limited hardware. For effectively separating strategy and language generation, we leverage action-utterance pairs (seen before).

For stability and efficiency purposes we applied cyclic LR scheduling, EMA loss tracking, layerwise LR decay, gradient checkpointing. A cyclic learning rate schedule that iterates through multiple warmup and decay phases. This approach helps the model escape local minima and explore a broader parameter space, potentially leading to better generalization and performance. EMA smoothing on the training loss provides a stable and noise-resistant indicator of long-term training progress. This aids in distinguishing genuine improvements from short-term fluctuations. 

Then, we used gradient checkpointing to reduce memory usage by recomputing certain activations during backpropagation. By combining gradient checkpointing and mixed precision training we were able to use longer sequence lengths and larger batch sizes within our hardware constraints. Specifically, we achieved an effective batch size of 32 with a per-device batch size of 16 and a sequence length of 512 tokens.

Finally, we implemented a layerwise learning rate decay which progressively lowers learning rates to lower (earlier) layers of the model, while higher (later) layers receive higher learning rates. This strategy preserves foundational language capabilities in the lower layers while allowing the upper layers to adapt more aggressively to negotiation-specific tasks.

\subsection{Results of Fine-Tuning}
We successfully fine-tuned three role-specialized negotiation agents:
\begin{itemize}
    \item \textbf{Buyer Specialist:} Optimized to excel in the buyer role with targeted strategies.
    \item \textbf{Seller Specialist:} Optimized to excel in the seller role with targeted strategies.
    \item \textbf{Generalist Negotiator:} Capable of adapting dynamically to both buyer and seller roles.
\end{itemize}

Training logs and Tensorboard metrics demonstrated:
\begin{itemize}
    \item Stable GPU memory usage (peak: 22GB, stable: 3.56GB).
    \item Gradual and consistent loss reduction across all role-specific models.
    \item Final validation losses converged around 4.05, with the generalist model achieving a slightly lower EMA loss of 4.49.
\end{itemize}

\begin{figure}[h!]
    \centering
    \includegraphics[width=0.45\linewidth]{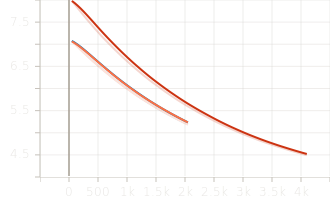}
    \includegraphics[width=0.45\linewidth]{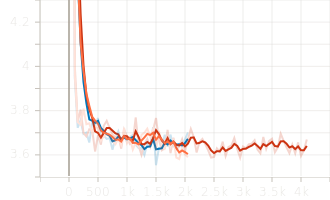}
    \caption{Training Loss Metrics. Left: EMA-smoothed loss. Right: Raw loss curve. Both plots demonstrate consistent loss convergence across 4k steps.}
    \label{fig:loss_plots}
\end{figure}

\begin{figure}[h!]
    \centering
    \includegraphics[width=0.6\linewidth]{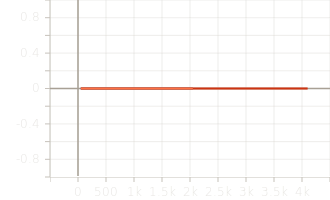}
    \caption{Steps Since Improvement. Minimal fluctuation indicates stable model convergence throughout training.}
    \label{fig:steps_since_improvement}
\end{figure}

\begin{figure}[h!]
    \centering
    \includegraphics[width=0.45\linewidth]{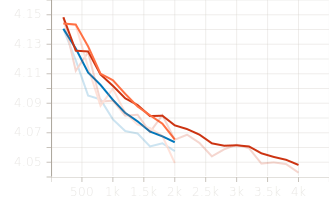}
    \includegraphics[width=0.45\linewidth]{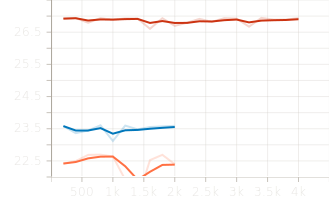}
    \caption{Evaluation Loss and Throughput Metrics. Left: Validation loss trends stabilize around 4.05. Right: Evaluation throughput demonstrates consistent samples per second.}
    \label{fig:eval_metrics}
\end{figure}

\begin{figure}[h!]
    \centering
    \includegraphics[width=0.45\linewidth]{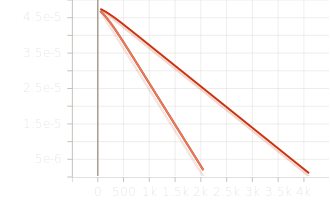}
    \includegraphics[width=0.45\linewidth]{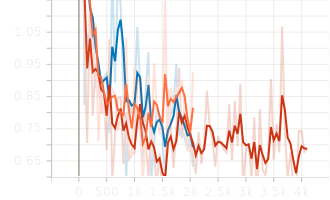}
    \caption{Learning Rate and Gradient Behavior. Left: Layerwise cyclic learning rate decay. Right: Gradient norms demonstrate stable backpropagation dynamics.}
    \label{fig:learning_rate_grad}
\end{figure}


\section{Experiments}
Using the techniques described earlier, we evaluated 21 models across 52 model combinations to explore their negotiation capabilities. This section covers our experimental setup, metrics, detailed analysis, and insights into fine-tuned behaviors, Chain of Thought (CoT), model size comparisons, and attention probing.

\subsection{Setup}
We designed our framework for testing as follows. First, 30 scenarios from the Craiglist Bargaining Dataset are randomly selected from the test split. These scenarios are used across all model combinations. 

For each scenario, we create buyer and seller knowledge bases which include information regarding the role of the agent, the posting's title, description, category, and listing price, and the agent's target price. 

Then, given a model combination (a buyer model and a seller model), buyer and seller knowledge bases, and a clean conversation history, a negotiation runner module orchestrates the interaction between the buyer and seller agent. A negotiation run takes a maximum of 15 turns where in each turn, a model (buyer or agent) generates an \textit{action-utterance} pair or an \textit{action-utterance-reasoning tuple} given a conversation history. Relevant information like prices and actions are recorded in each turn. At the end of a turn, the conversation history is updated with the \textit{utterance} generated.

After the negotiation concludes, either through an acceptance, rejection, or reaching the maximum turn limit, we record the latest conversation history and calculate a series of metrics detailed in \ref{subsec:metrics}. A tester module runs the negotiation runner for all 30 scenarios across any given model combination and stores the results in a csv file. 

In our experiments we evaluate the performance of the following 21 models: 
\begin{itemize}
    \item \textbf{Baseline Models}
    \begin{itemize}
        \item Llama 3.2 (3B, 8B, 70B) as buyers
        \item Llama 3.2 (3B, 8B, 70B) as sellers
    \end{itemize}

    \item \textbf{Baseline Models w/ COT} 
    \begin{itemize}
        \item Llama 3.2 (3B, 8B, 70B) as buyers with CoT 
        \item Llama 3.2 (3B, 8B, 70B) as sellers with CoT
    \end{itemize}
    
    \item \textbf{Personality Models} 
    \begin{itemize}
        \item Llama 3.2 (3B) as buyers with aggressive, fair, and passive passive personalities.
        \item Llama 3.2 (3B) as sellers with aggressive, fair, and passive passive personalities
    \end{itemize}

    \item \textbf{Fine-tuned Models} 
    \begin{itemize}
        \item Llama 3.2 (3B) Buyer Specialist
        \item Llama 3.2 (3B) Seller Specialist
        \item Llama 3.2 (3B) Generalist
    \end{itemize}
\end{itemize}

\subsection{Metrics}
\label{subsec:metrics}
We collect the following metrics after performing a negotiation run between a buyer and seller. Additionally, we detail some aggregated metrics here.
\\ \\
\noindent
\textbf{Aggreement Rate } Measures the proportion of negotiations that ended in an acceptance.
\begin{align*}
    \text{Agreement Rate} = \frac{\text{\# Accepted Negotiations}}{\text{\# Negotiations}}
\end{align*}
\noindent
\textbf{Dialogue Length } Counts the number of turns taken in a negotiation run. $T$ is the set of turns.
\begin{align*}
    \text{Dialogue Length} = \lvert T \rvert
\end{align*}
\noindent
\textbf{Fairness } Measures how equitable the negotiation outcome was for both parties relative to the midpoint of the agent target prices. A fairness of $1$ means the negotiation outcome was perfectly fair while $0$ means maximally unfair.
\begin{align*}
    \text{Fairness} = 1 - 2 \cdot \frac{\lvert \$_{\text{accepted}} -  \frac{(\$_{\text{seller target}} + \$_{\text{buyer target}})}{2} \rvert}{\$_{\text{seller target}} - \$_{\text{seller target}}}
\end{align*}
\noindent
\textbf{Aggressiveness } Measures the magnitude of deviation between the accepted and listing prices.
\begin{align*}
    \text{Aggressiveness} = \frac{\lvert \$_{\text{accepted}} - \$_{\text{listing}} \rvert}{\$_{\text{listing}}}
\end{align*}
\noindent
\textbf{Bias } Measures the relative deviation from the seller's target price. A bias of $-1$ means the negotiation completely favors the seller. A bias of $+1$ means the negotiation completely favors the buyer. 
\begin{align*}
    \text{Bias} = 2 \cdot \frac{\lvert \$_{\text{seller target}} - \$_{\text{accepted}} \rvert}{\$_{\text{seller target}} - \$_{\text{buyer target}}} - 1
\end{align*}
\noindent
\textbf{Bias Conditioned on Dialogue Length} Scales bias result by the dialogue length. We were interested to see if longer negotiations tend to reduce or amplify bias.  
\begin{align*}
    \text{Bias Cond Dialogue Length} = \frac{\text{Bias}}{\text{Dialogue Length}}
\end{align*}
\noindent
\textbf{Concession Rate} Measures how flexible price proposals were in the negotiation. Higher values indicate a greater flexibility while lower values indicate a stagnant price action. 
\begin{align*}
    \text{Concession Rate} = \frac{\sum \lvert \$_{\text{current price act}} - \$_{\text{previous price act}}\rvert}{\text{Dialogue Length}}
\end{align*}
\noindent
\textbf{Relative Efficiency} Looks at fairness per turn. Here, we were intersted to see how quickly a negotiation reaches a balanced (ideally) outcome.
\begin{align*}
    \text{Relative Efficiency} = \frac{\text{Fairness}}{\text{Dialogue Length}}
\end{align*}
\noindent
\textbf{Probing Ratio} Measures the number of inquiries performed in the negotiation.
\begin{align*}
    \text{Probing Ratio} = \frac{\text{\# \textit{inquiry} acts}}{\text{Dialogue Length}}
\end{align*}


\subsection{Personality Analysis}
Models were tested in same-role and mixed-role personality configurations:
\begin{itemize}
    \item \textbf{Same-Role Combinations}: e.g., Aggressive Buyer + Aggressive Seller.
    \item \textbf{Mixed-Role Combinations}: e.g., Aggressive Buyer + Fair Seller.
\end{itemize}

\subsubsection{Agreement Rates}
Table~\ref{tab:agreement_rates} shows the agreement rates across all personality pairings. Models with a passive seller and aggressive buyer achieved the highest success.

\begin{table}[h!]
\centering
\begin{tabular}{l|c}
\hline
\textbf{Personality Combination} & \textbf{Agree Rate} \\
\hline
Aggressive Buyer + Fair Seller   & 0.85 \\
Aggressive Buyer + Passive Seller & 0.72 \\
Fair Buyer + Aggressive Seller   & 0.78 \\
Passive Buyer + Passive Seller   & 0.64 \\
Passive Buyer + Fair Seller      & 0.80 \\
\hline
\end{tabular}
\caption{Agreement rates for personality combinations.}
\label{tab:agreement_rates}
\end{table}

\begin{figure}[h]
    \centering
    \includegraphics[width=1\linewidth]{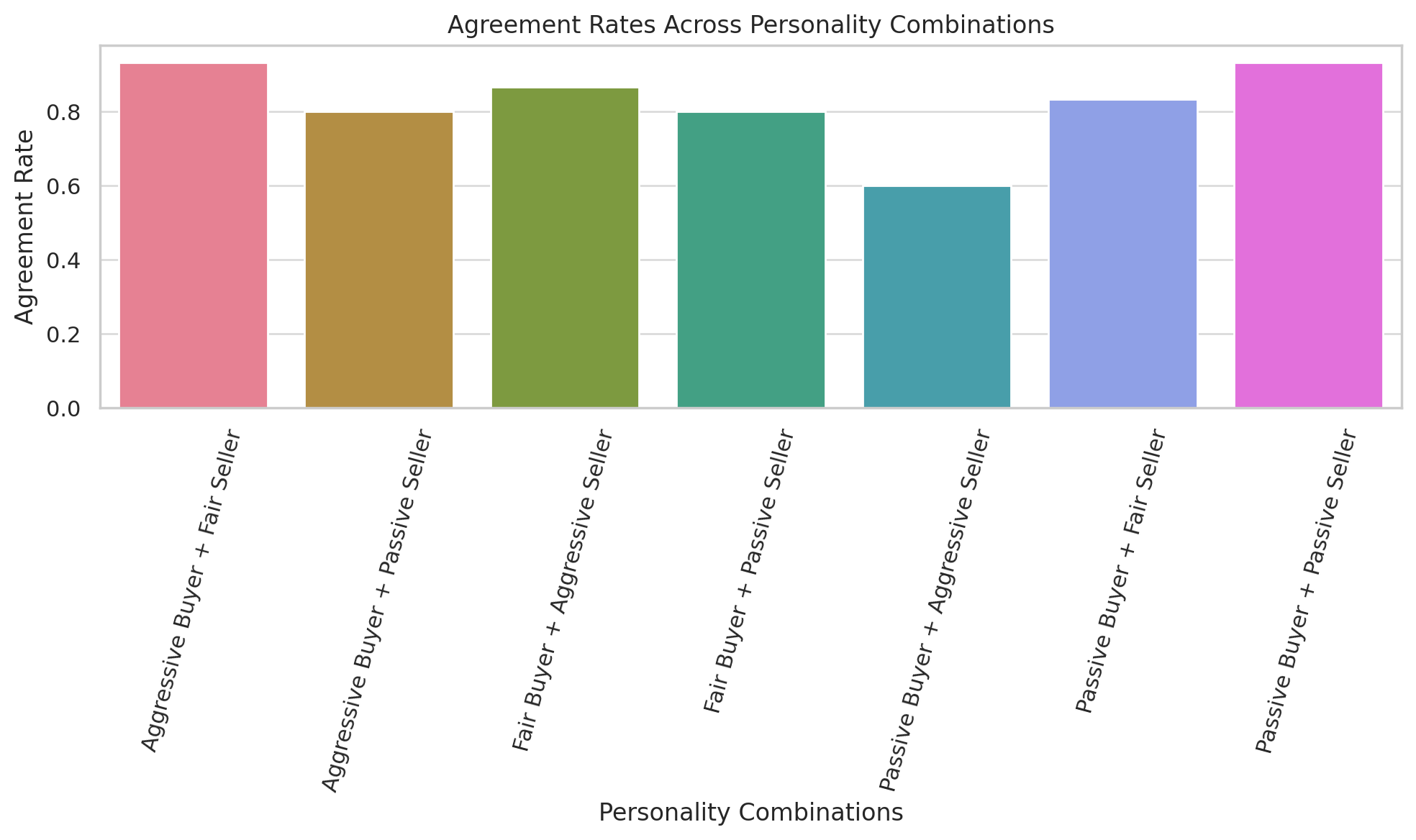}
    \caption{Agreement Rates Across Personality Combinations. Aggressive buyers paired with fair sellers achieved the highest success, while passive combinations exhibited moderate agreement rates.}
    \label{fig:agreement_rates}
\end{figure}

\subsubsection{Bias Toward Buyer/Seller}
We find that aggressive buyers dominated interactions with sellers conceding more. Passive sellers showed the highest buyer biases.

\begin{figure}[h]
    \centering
    \includegraphics[width=1\linewidth]{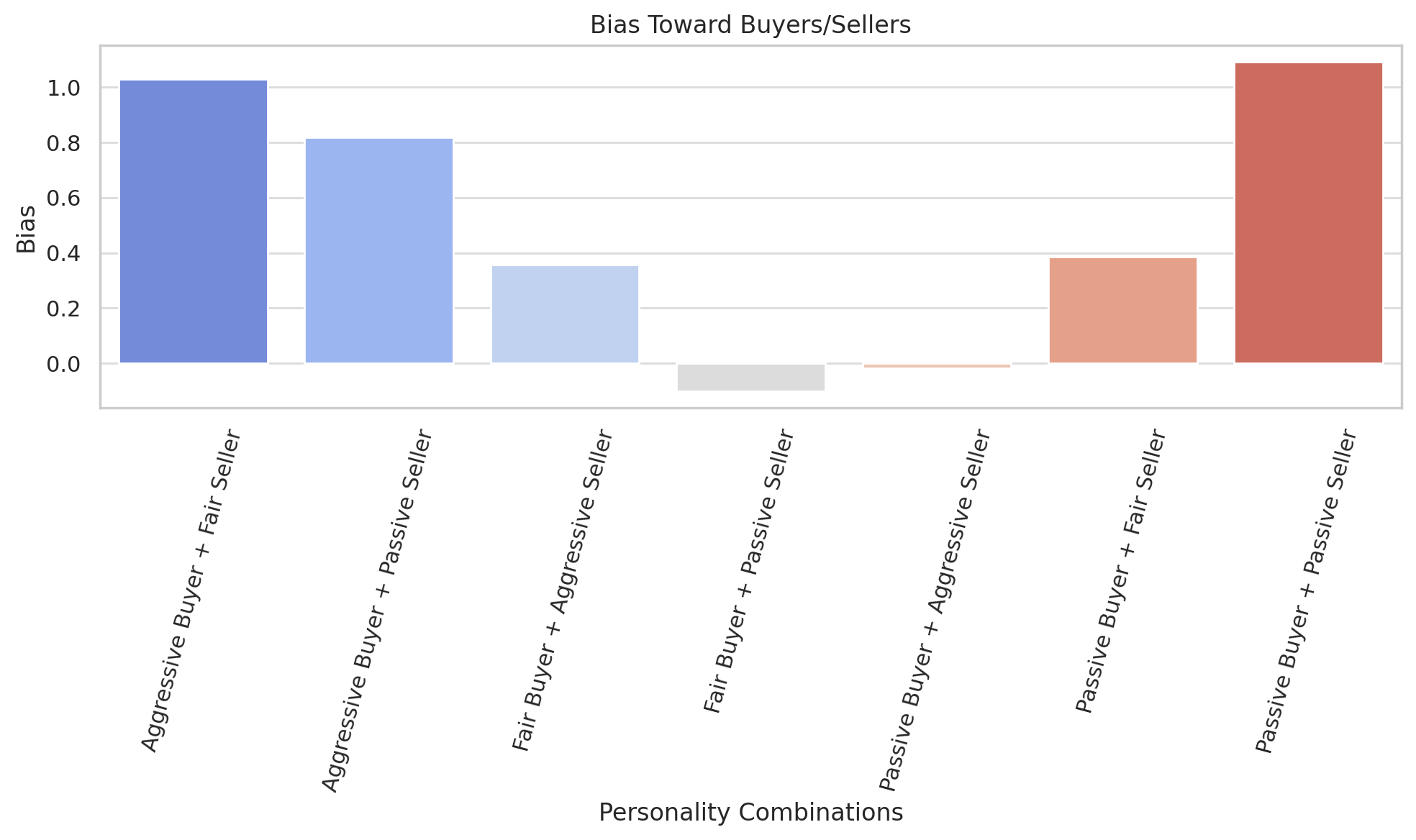}
    \caption{Bias Toward Buyers/Sellers Across Personality Combinations. Aggressive buyers dominate negotiations, while passive sellers exhibit the highest buyer biases.}
    \label{fig:bias_scores}
\end{figure}

\subsubsection{Fairness Scores}
Results in Table~\ref{tab:fairness_scores} show aggressive buyers compromised fairness the most.

\begin{table}[h!]
\centering
\begin{tabular}{l|c}
\hline
\textbf{Personality Combination} & \textbf{Fairness} \\
\hline
Aggressive Buyer + Fair Seller   & -0.35 \\
Fair Buyer + Passive Seller      & 0.10  \\
Passive Buyer + Passive Seller   & -0.50 \\
\hline
\end{tabular}
\caption{Fairness scores across personality combinations.}
\label{tab:fairness_scores}
\end{table}

\begin{figure}[h]
    \centering
    \includegraphics[width=1\linewidth]{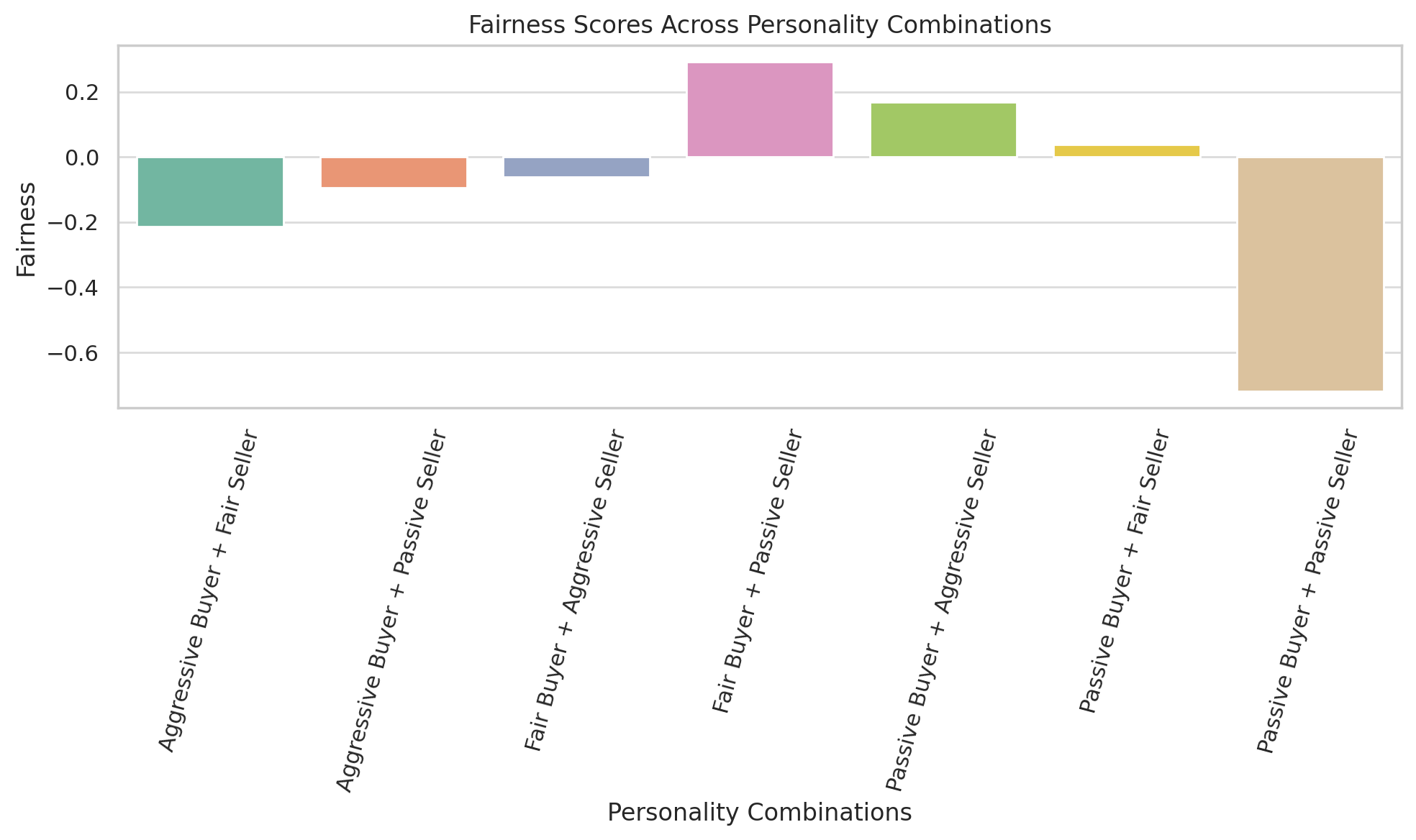}
    \caption{Fairness Scores Across Personality Combinations. Aggressive buyers compromise fairness, while fair pairings show balanced outcomes. Passive sellers show extreme deviations.}
    \label{fig:fairness_scores}
\end{figure}

\subsubsection{Dialogue Lengths and Price Progression}
Aggressive combinations yielded shorter, intense negotiations, while fair pairings led to balanced turn exchanges. Passive pairings resulted in longer sequences with extended price deliberations.

\begin{figure}[h]
    \centering
    \includegraphics[width=1\linewidth]{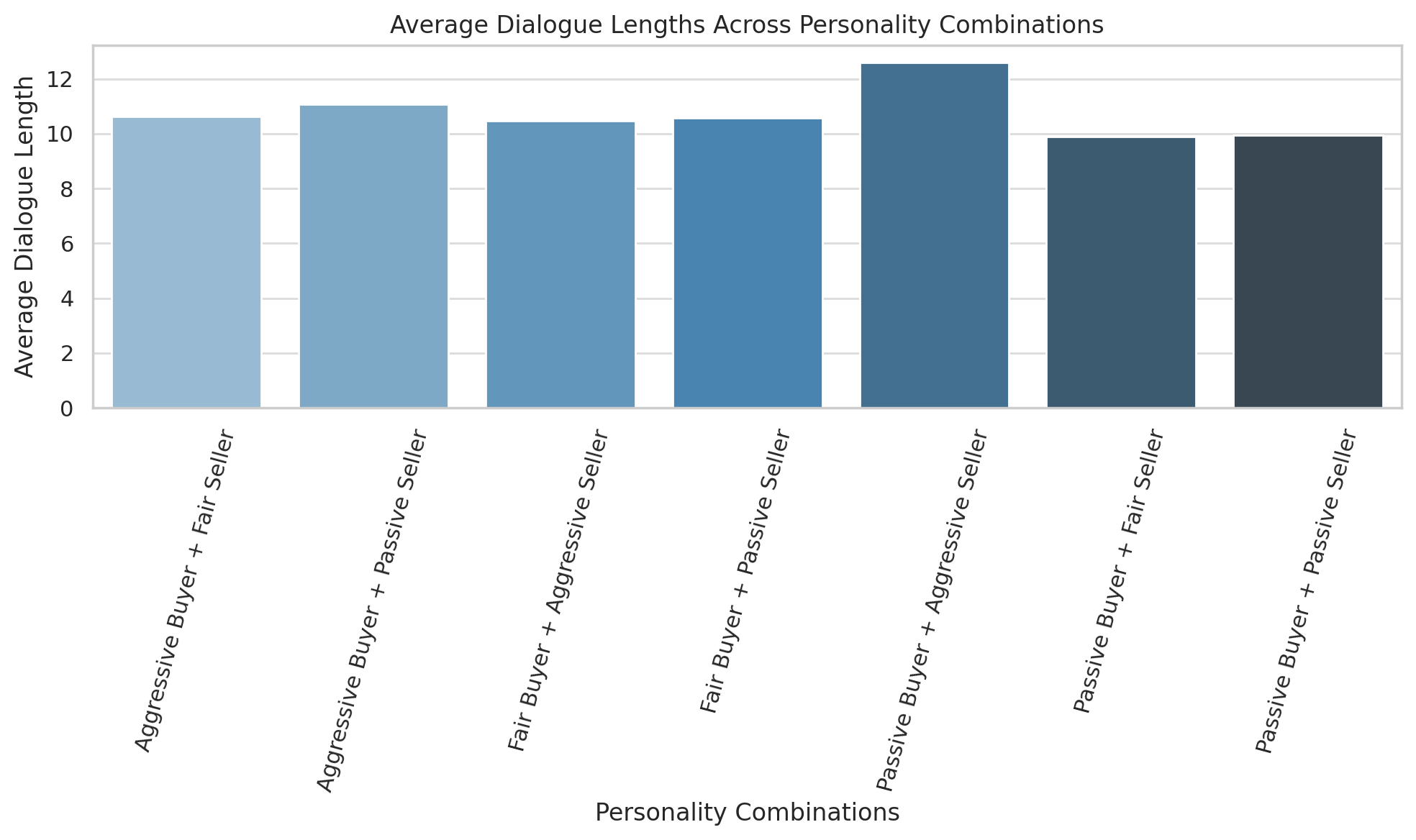}
    \caption{Average Dialogue Lengths Across Personality Combinations. Aggressive combinations show shorter dialogues, while fair combinations tend to stabilize conversation lengths.}
    \label{fig:dialogue_lengths}
\end{figure}

\begin{figure}[h]
    \centering
    \includegraphics[width=1\linewidth]{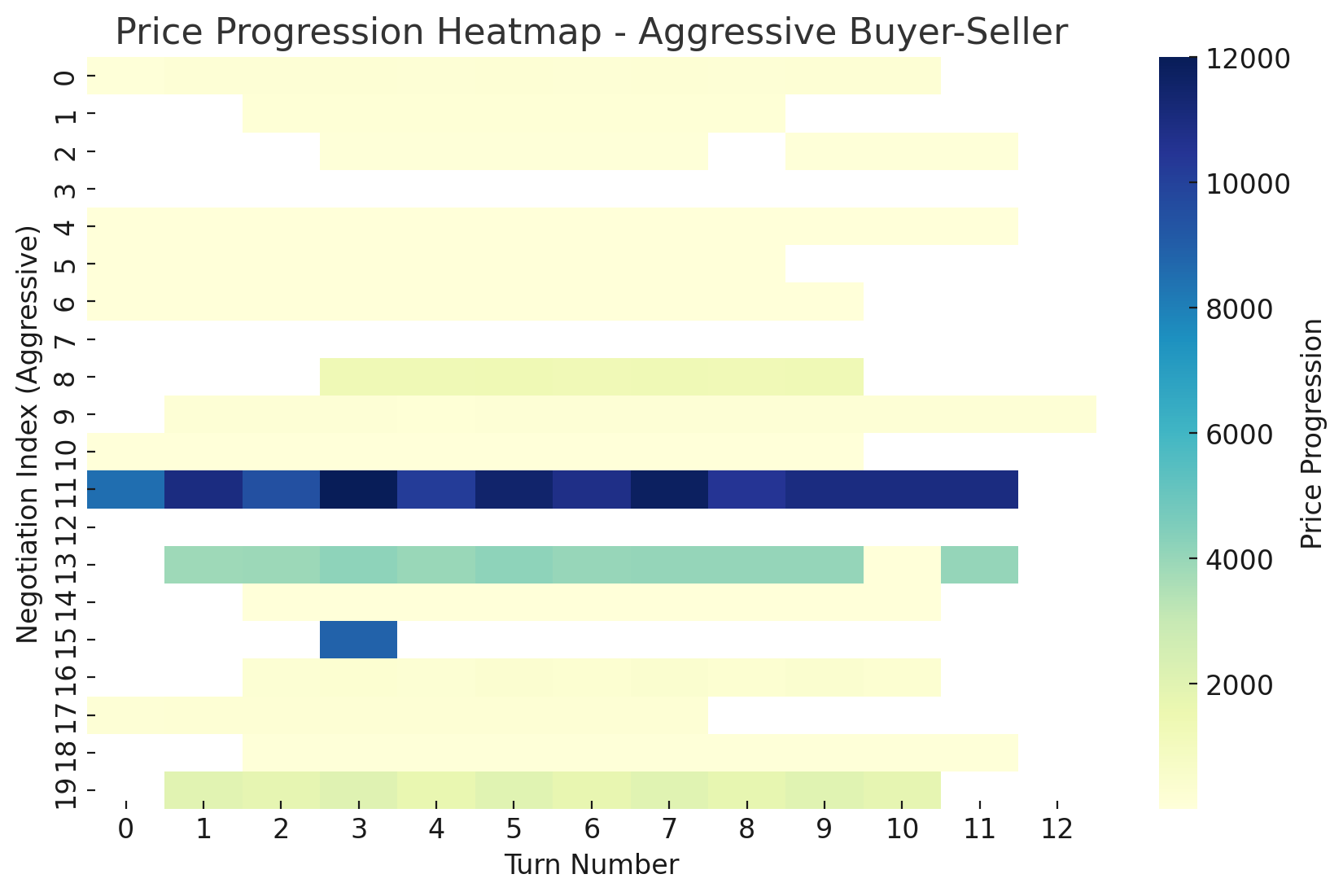}
    \caption{Price Progression Heatmap for Aggressive Buyer-Seller Combinations. Aggressive interactions lead to steep price adjustments early in negotiations.}
    \label{fig:price_progression_aggressive}
\end{figure}

\begin{figure}[h]
    \centering
    \includegraphics[width=1\linewidth]{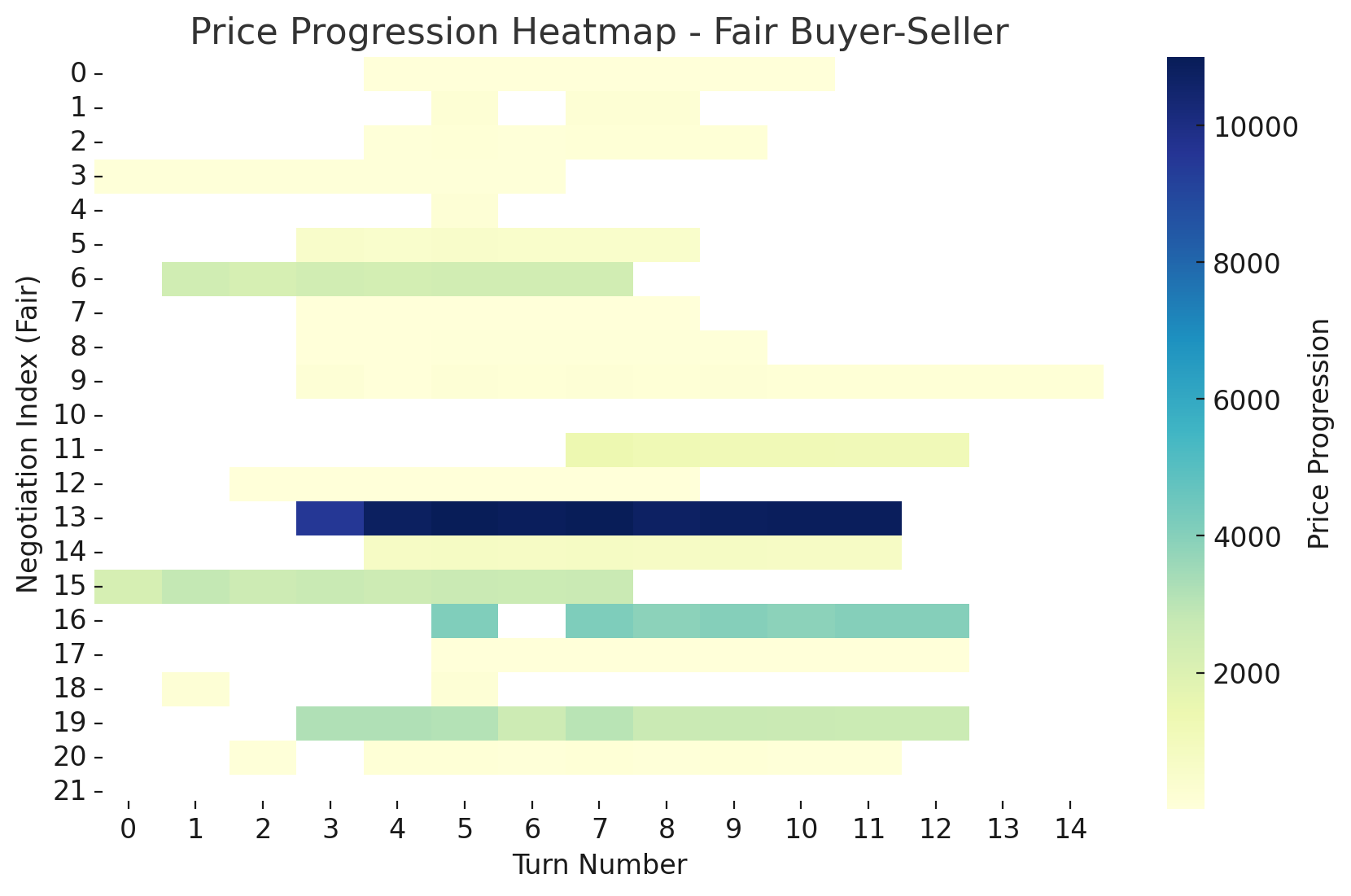}
    \caption{Price Progression Heatmap for Fair Buyer-Seller Combinations. Fair combinations maintain moderate price adjustments, demonstrating balance.}
    \label{fig:price_progression_fair}
\end{figure}

\begin{figure}[h]
    \centering
    \includegraphics[width=1\linewidth]{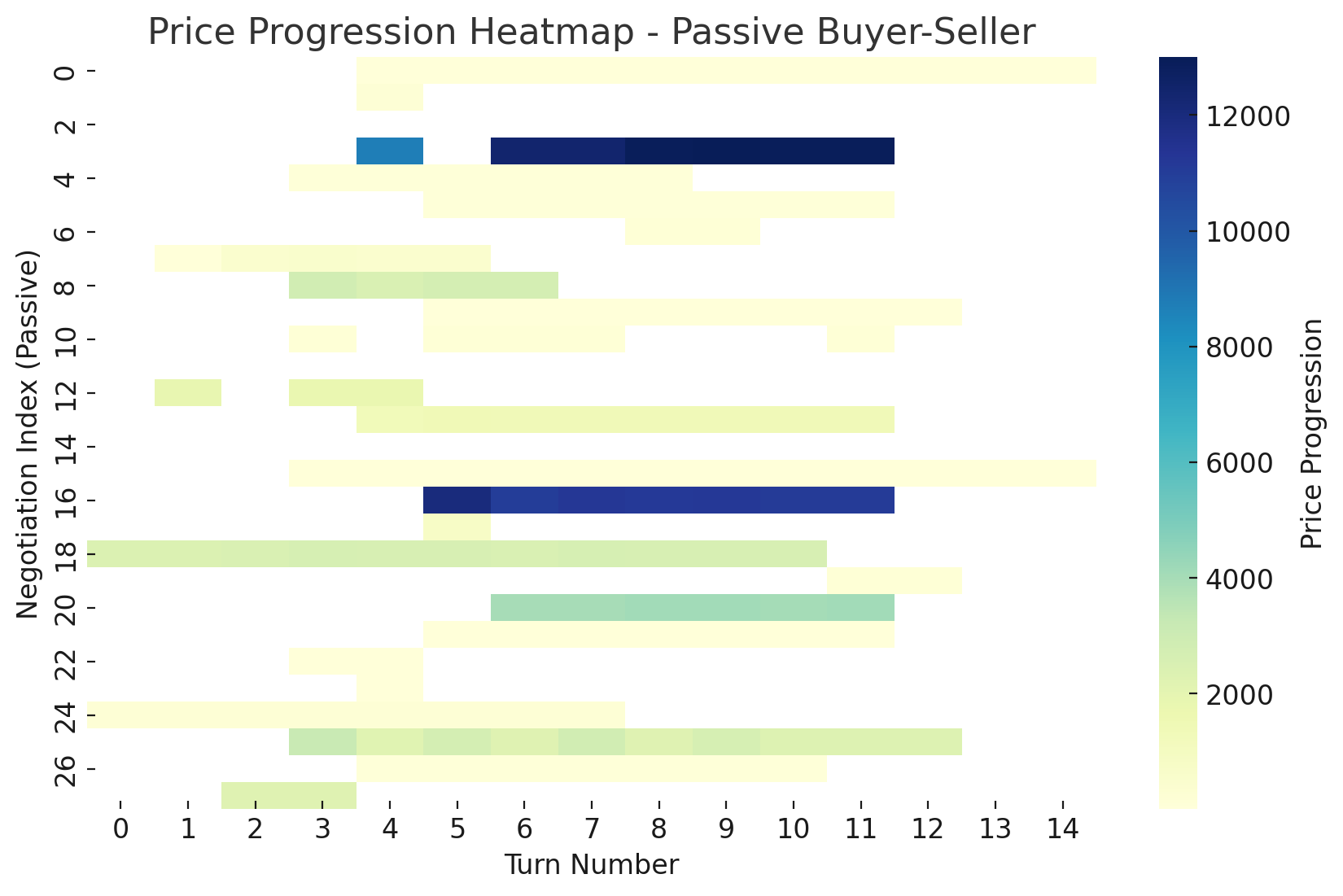}
    \caption{Price Progression Heatmap for Passive Buyer-Seller Combinations. Passive interactions show gradual price changes with prolonged negotiations.}
    \label{fig:price_progression_passive}
\end{figure}

\begin{figure}[h]
    \centering
    \includegraphics[width=1\linewidth]{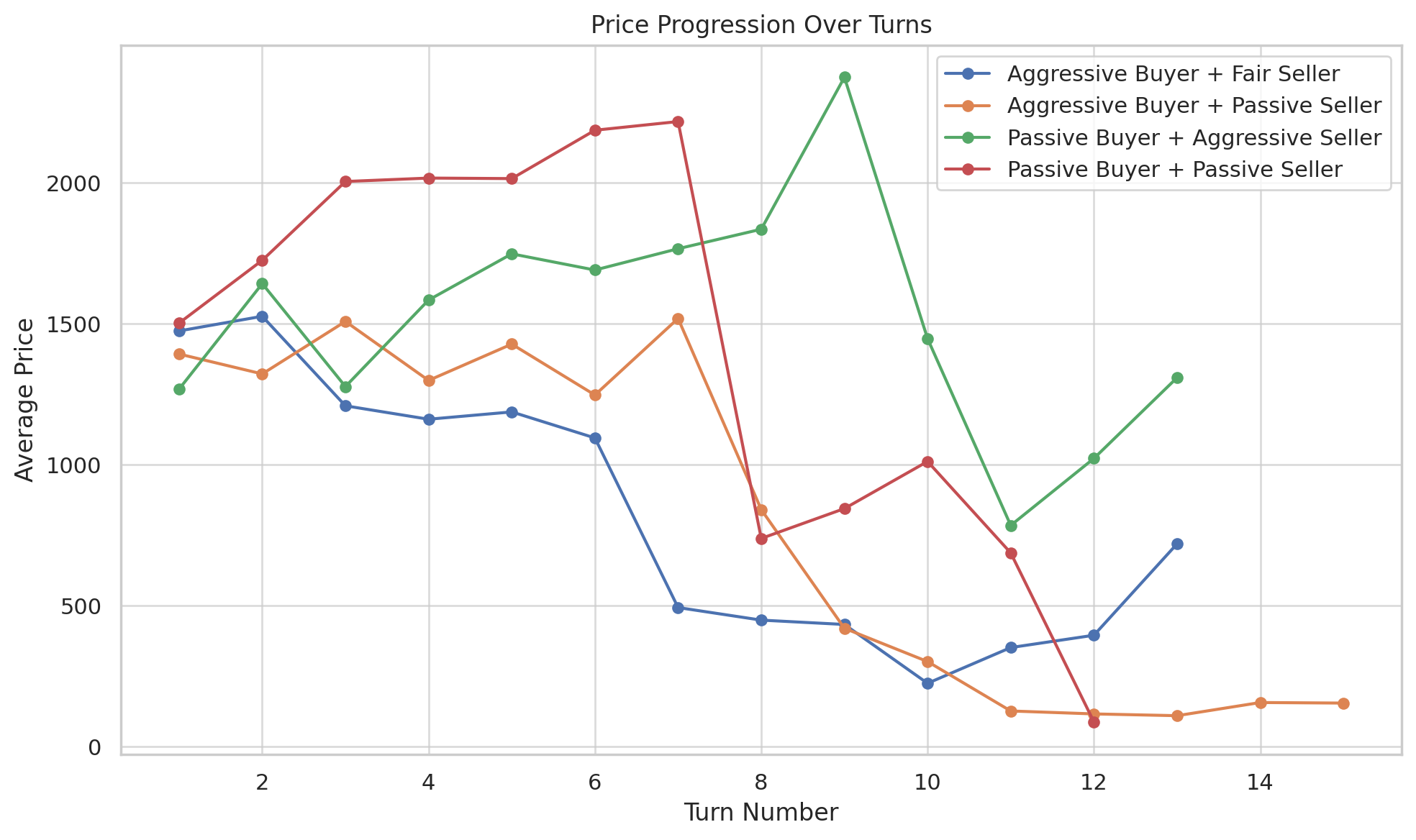}
    \caption{Price Progression Trends Over Turns. Comparative trends reveal aggressive buyers converge quickly, while passive negotiations take longer to stabilize.}
    \label{fig:price_progression_over_turns}
\end{figure}

\subsubsection{Action Distributions}
\begin{figure}[h]
    \centering
    \includegraphics[width=1\linewidth]{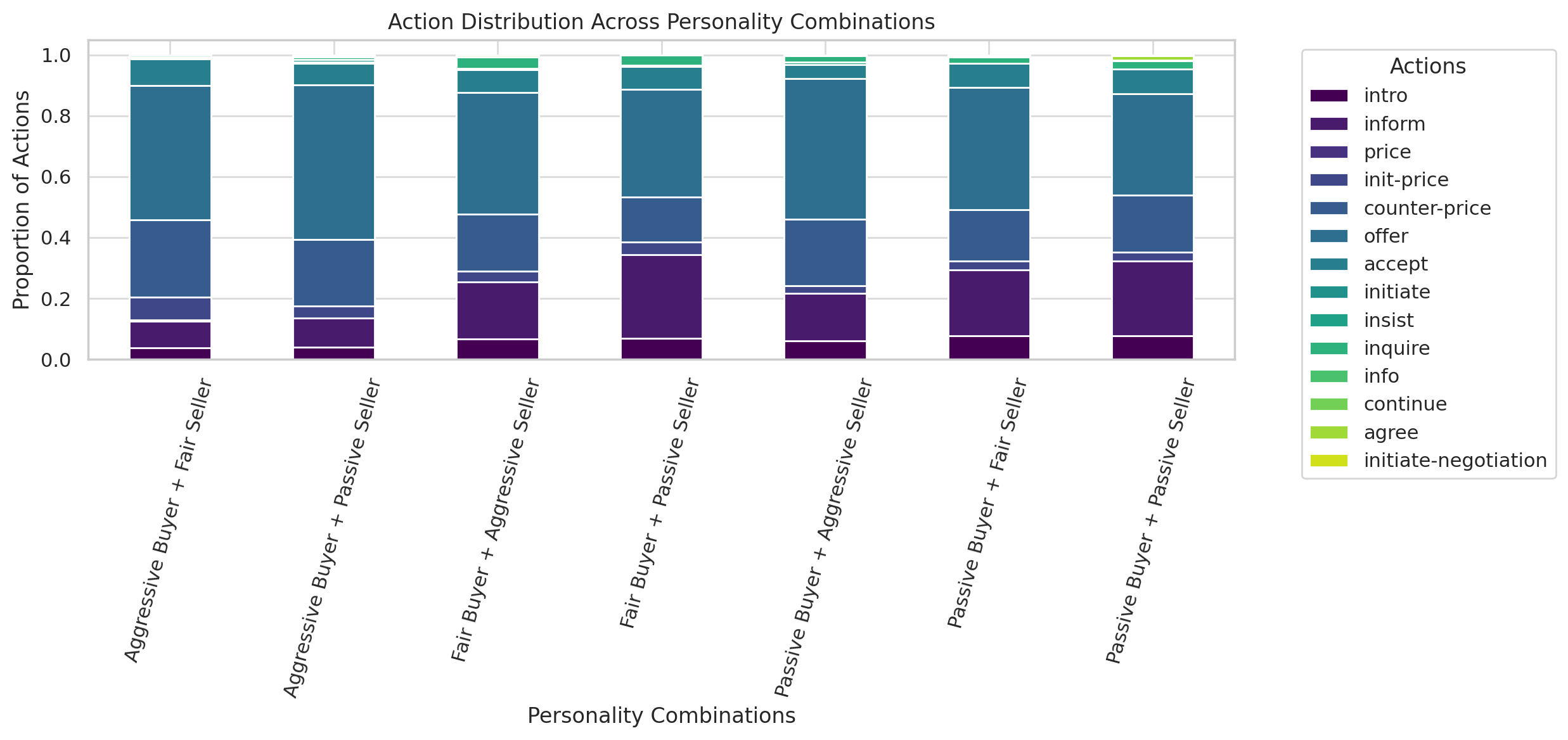}
    \caption{Action Distribution Across Personality Combinations. Aggressive agents prioritize direct negotiation tactics, while passive agents focus on informational acts.}
    \label{fig:action_distributions}
\end{figure}

Figure~\ref{fig:action_distributions} demonstrates personality-driven action preferences. Aggressive agents emphasize \textit{counter-price} and \textit{offer} actions, whereas passive agents prioritize \textit{inform} and avoid price counters.

\begin{figure}[h]
    \centering
    \includegraphics[width=1\linewidth]{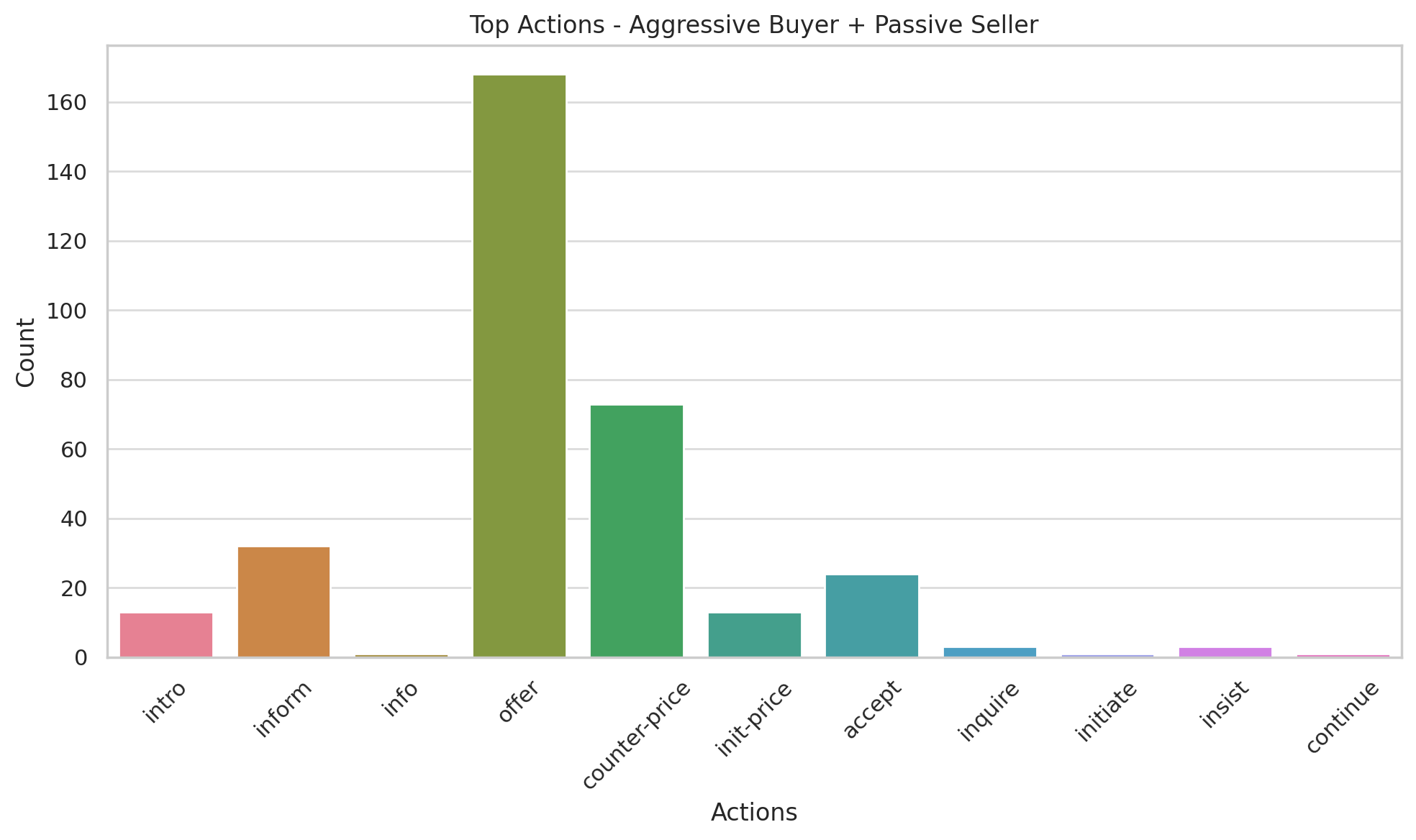}
    \caption{Top Actions for Aggressive Buyer + Passive Seller Scenario. Aggressive buyers dominate the negotiation with \textit{offer} and \textit{counter-price} acts.}
    \label{fig:top_actions_aggressive_passive}
\end{figure}

\subsubsection{Sequence Lengths}
\begin{figure}[h]
    \centering
    \includegraphics[width=1\linewidth]{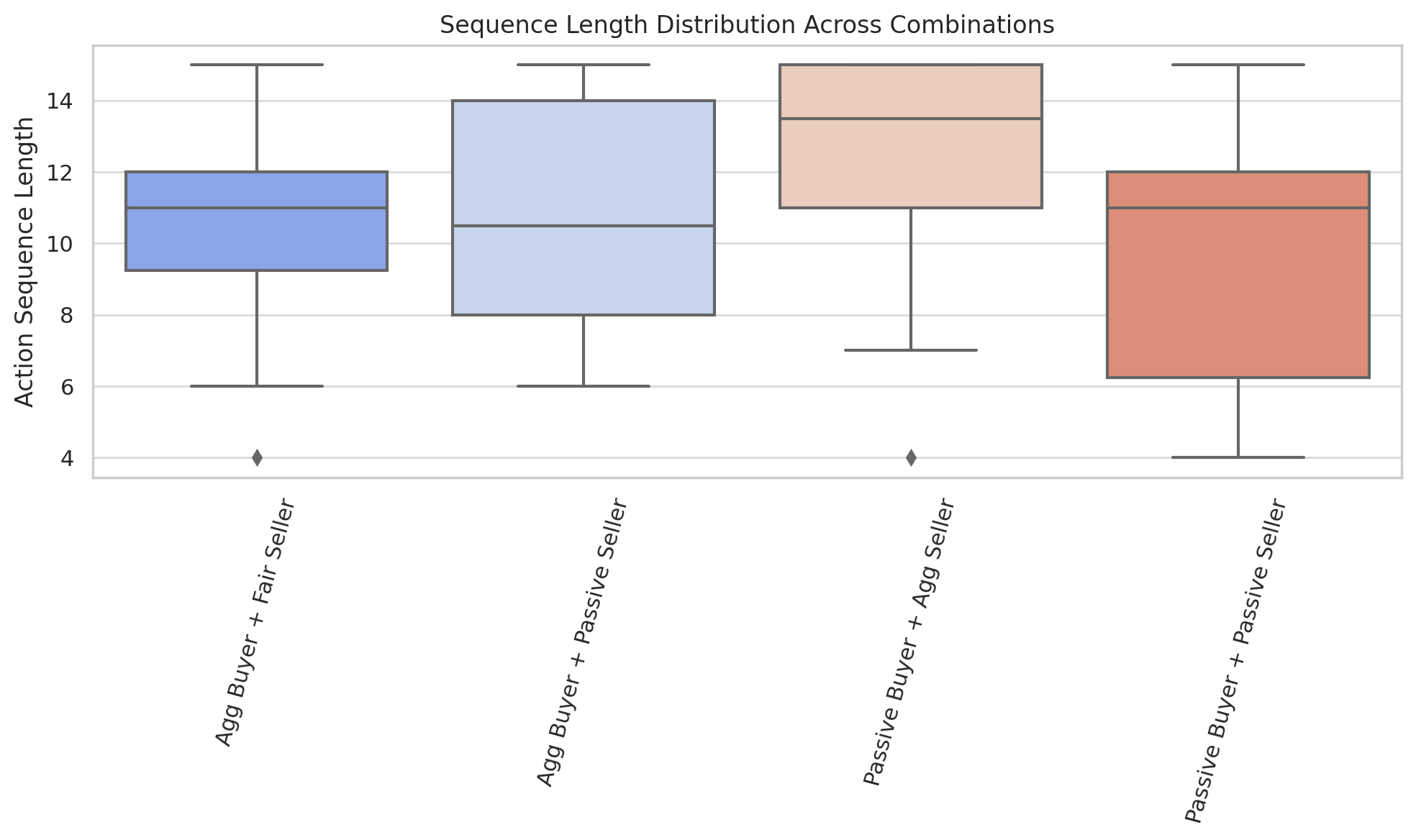}
    \caption{Action Sequence Length Distribution Across Personality Combinations. Passive combinations generate longer sequences, while aggressive pairs tend to resolve quickly.}
    \label{fig:sequence_lengths}
\end{figure}

Sequence lengths provide insights into negotiation dynamics. Figure~\ref{fig:sequence_lengths} shows that passive scenarios produce longer interactions, with aggressive combinations being concise.

\subsubsection{Key Takeaways}
The personality-based scenarios reveal some interesting trends:

\begin{itemize}
    \item \textbf{Agreement Rates}: Aggressive buyers paired with fair sellers achieved the highest agreement rates (Figure~\ref{fig:agreement_rates}), demonstrating that a mix of assertive and balanced strategies leads to successful outcomes. Passive pairings, while cooperative, show lower success due to prolonged deliberations.

    \item \textbf{Bias}: Aggressive buyers dominate negotiations, securing favorable outcomes for themselves, especially against passive sellers. Passive sellers exhibit the highest buyer bias (Figure~\ref{fig:bias_scores}), underscoring their tendency to concede.

    \item \textbf{Fairness}: Aggressive combinations compromise fairness the most, deviating significantly from midpoint agreements. Fair and balanced pairings achieve relatively equitable outcomes (Figure~\ref{fig:fairness_scores}).

    \item \textbf{Dialogue Lengths and Price Progression}: 
    \begin{itemize}
        \item \textit{Aggressive Combinations} result in shorter dialogues with steep price adjustments early in the negotiation (Figures~\ref{fig:dialogue_lengths}, \ref{fig:price_progression_aggressive}).
        \item \textit{Fair Combinations} balance dialogue length and price progression, promoting steady compromises (Figure~\ref{fig:price_progression_fair}).
        \item \textit{Passive Combinations} yield prolonged dialogues with gradual price movements, reflecting hesitancy in decision-making (Figure~\ref{fig:price_progression_passive}).
    \end{itemize}

    \item \textbf{Action Distributions}: Aggressive agents prioritize \textit{offer} and \textit{counter-price} actions, focusing on reaching their target quickly. Passive agents rely more on \textit{inform} and avoid assertive negotiation tactics (Figure~\ref{fig:action_distributions}).

    \item \textbf{Sequence Lengths}: Passive negotiations produce longer action sequences, while aggressive interactions resolve efficiently (Figure~\ref{fig:sequence_lengths}).
\end{itemize}

\noindent The intuition that personality traits should noticeably influence negotiation dynamics appears to hold. Aggressive strategies dominate in efficiency and bias toward the buyer but at the cost of fairness. Fair strategies strike a balance between outcomes and dialogue structure, while passive agents ensure cooperative, albeit slower, negotiations. These findings emphasize the importance of aligning negotiation goals with personality-driven behaviors to optimize outcomes.


\subsection{Size Comparison \& CoT Analysis}
In this section we analyze how model scale and reasoning influence the performance of LLMs on negotiation tasks. We test all combinations of the following models (36 in total). 
\begin{itemize}
    \item Llama 3.2 (3B, 8B, 70B) as buyers
    \item Llama 3.2 (3B, 8B, 70B) as sellers
    \item Llama 3.2 (3B, 8B, 70B) as buyers with CoT 
    \item Llama 3.2 (3B, 8B, 70B) as sellers with CoT
\end{itemize}

\subsubsection{Agreement Rates}
We observe that larger models consistently show higher agreement rates. Smaller models vary more in their agreement rates and are generally lower. Because larger models have more parameters, they are able to learn more nuanced representations of the negotiation state and hence adhere better to the goal of closing a deal. Moreover, since they show higher agreement rates in combinations with smaller models, this seems to indicate that larger models can better adapt to / counter bolder or aggressive proposals. Interestingly, we find that combinations involving 8B parameter models yield significantly lower agreement rates.  
\begin{figure}[h]
    \centering
    \includegraphics[width=1\linewidth]{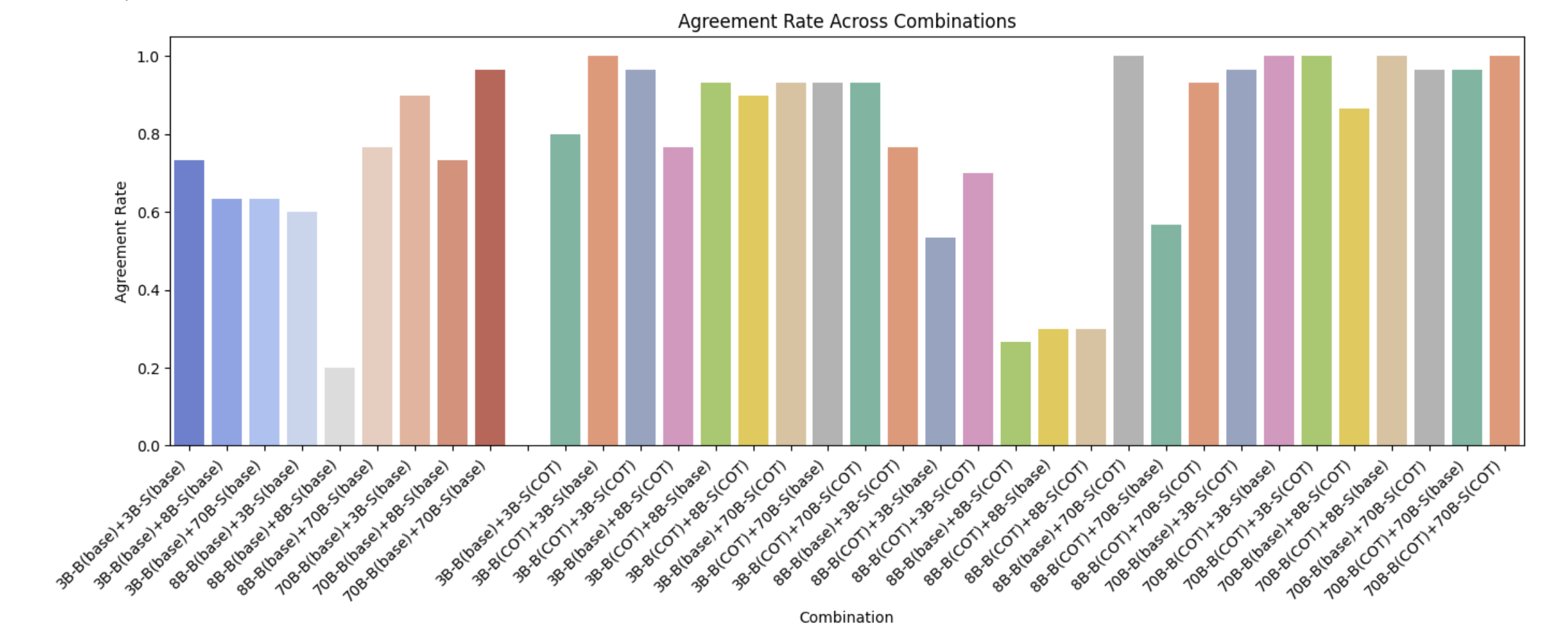}
    \caption{Agreement Rate across combinations. Shows the number of negotiations closed / ended in an accept act for each combination}
    \label{fig:agree}
\end{figure}

\subsubsection{Average Bias}
\begin{figure}[h]
    \centering
    \includegraphics[width=1\linewidth]{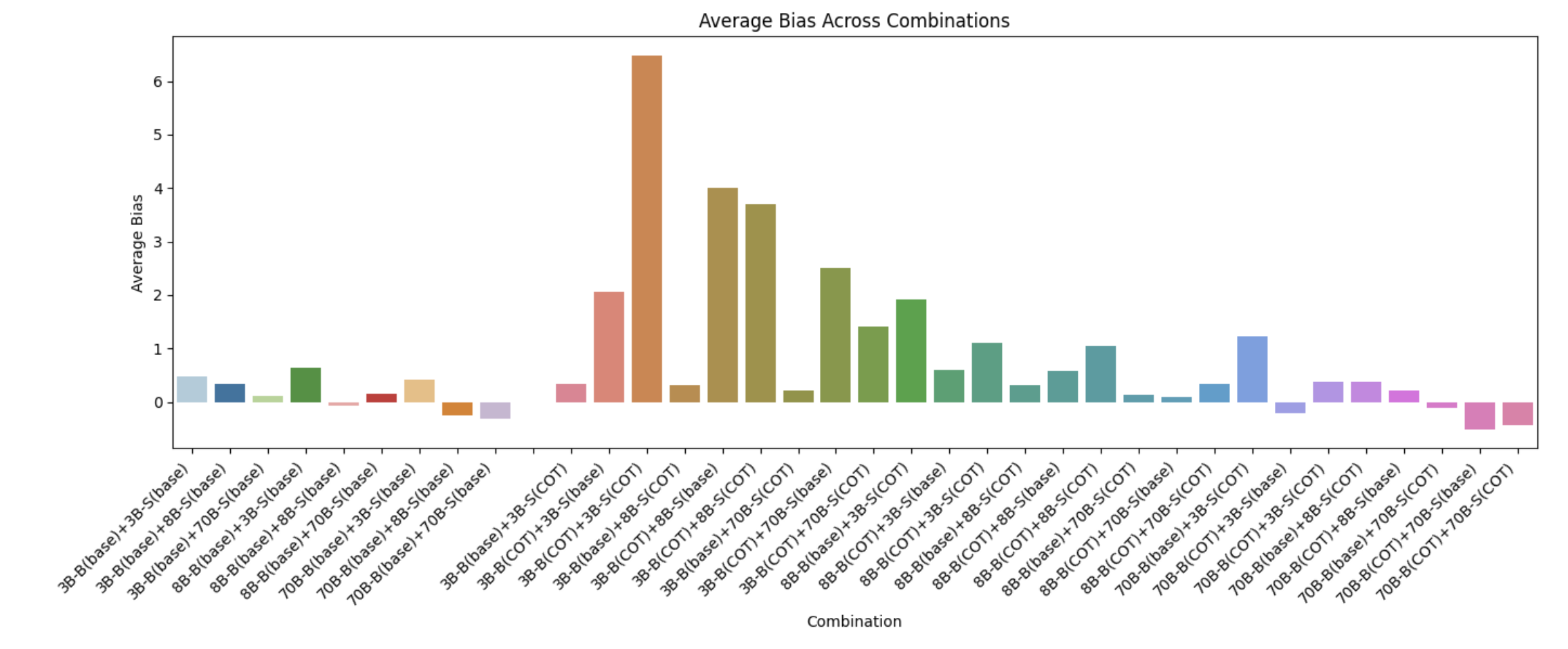}
    \caption{Average bias across combinations. A positive value indicates that the negotiation outcome was biased towards the buyer. 0 means no bias and any value below 0 indicates that the negotiation outcome was biased towards the seller.}
    \label{fig:avg_bias_size_cot}
\end{figure}

As seen in \ref{fig:avg_bias_size_cot}, smaller models tend to be more biased towards the buyer while larger models tend to be more biased towards the seller, although to a much lesser degree than the smaller models. We also find that combinations involving smaller models with CoT show very large biases toward the buyer.

\subsubsection{Average Fairness}
We observe that larger models are generally fairer than smaller models. Combinations with smaller models and CoT seem to yield extremely unfair negotiation outcomes while CoT with larger models seems to improve fairness.
\begin{figure}[h]
    \centering
    \includegraphics[width=1\linewidth]{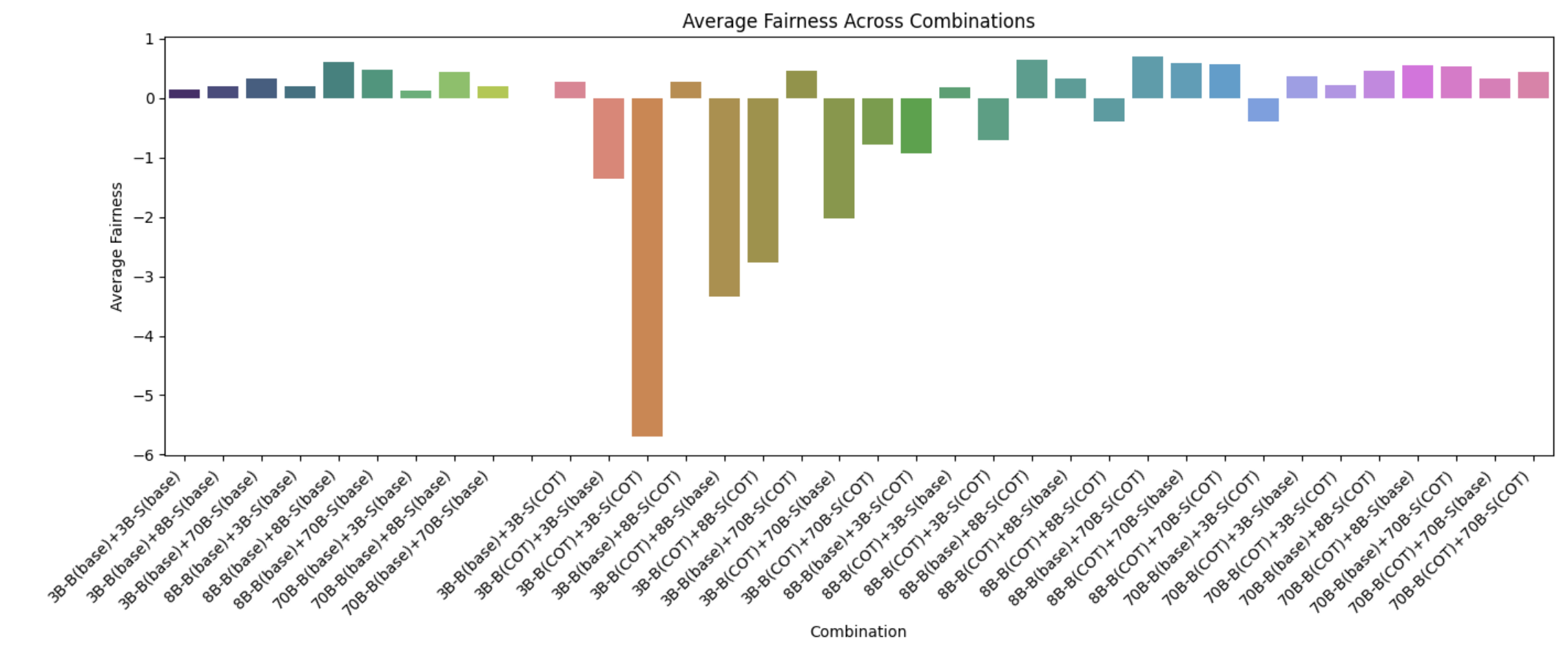}
    \caption{Average Fairness across all combinations}
    \label{fig:avg_fairness_size_cot}
\end{figure}

\subsubsection{Average Aggressiveness}
\begin{figure}[h]
    \centering
    \includegraphics[width=1\linewidth]{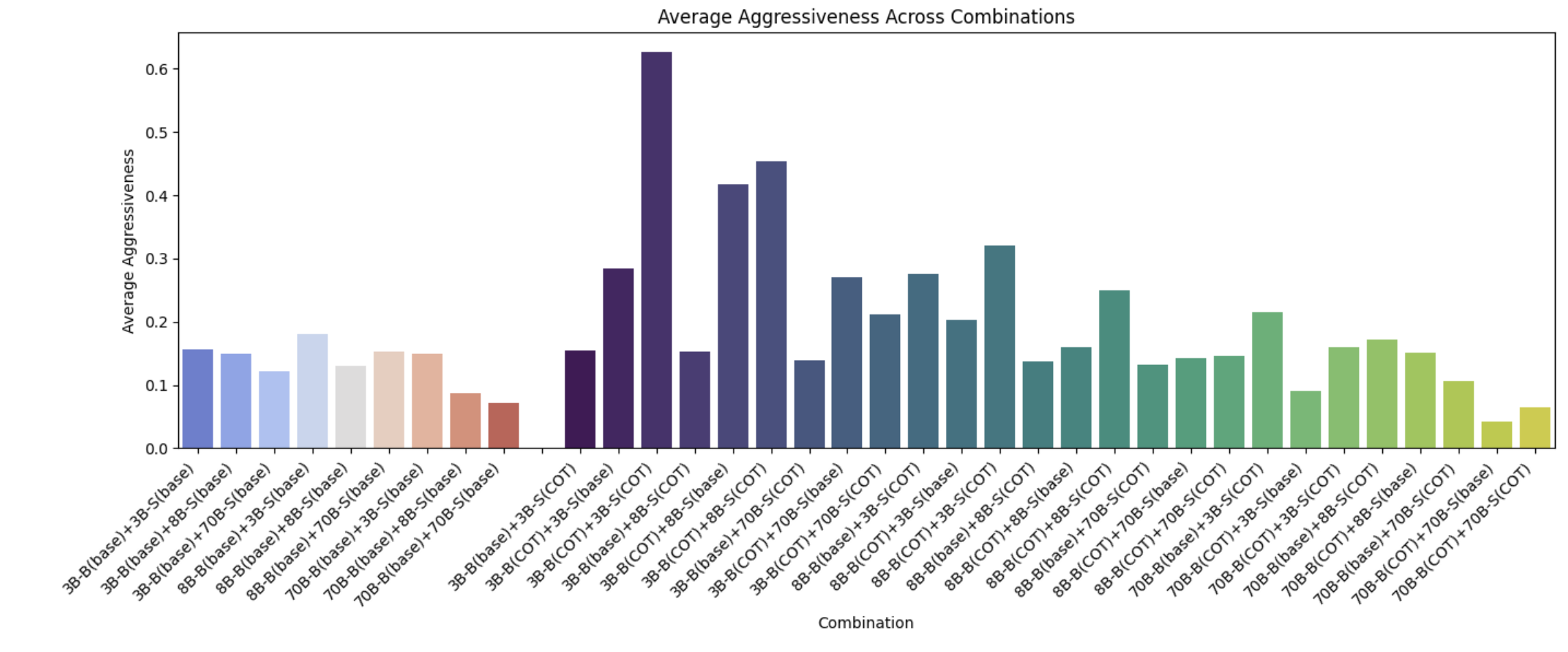}
    \caption{Average Aggressiveness across all combinations}
    \label{fig:avg_agg}
\end{figure}

Results in Fig \ref{fig:avg_agg} show that larger models exhibit less aggressiveness. Combinations involving CoT are generally more aggressive with effects magnified for combinations involving smaller models. 

\subsubsection{Average Dialogue Length}
We observe that larger models tend to generate shorter dialogues that smaller models. We again interestingly find here that combinations involving the 8B parameter model generally produce longer dialogues especially when using CoT.
\begin{figure}[h]
    \centering
    \includegraphics[width=1\linewidth]{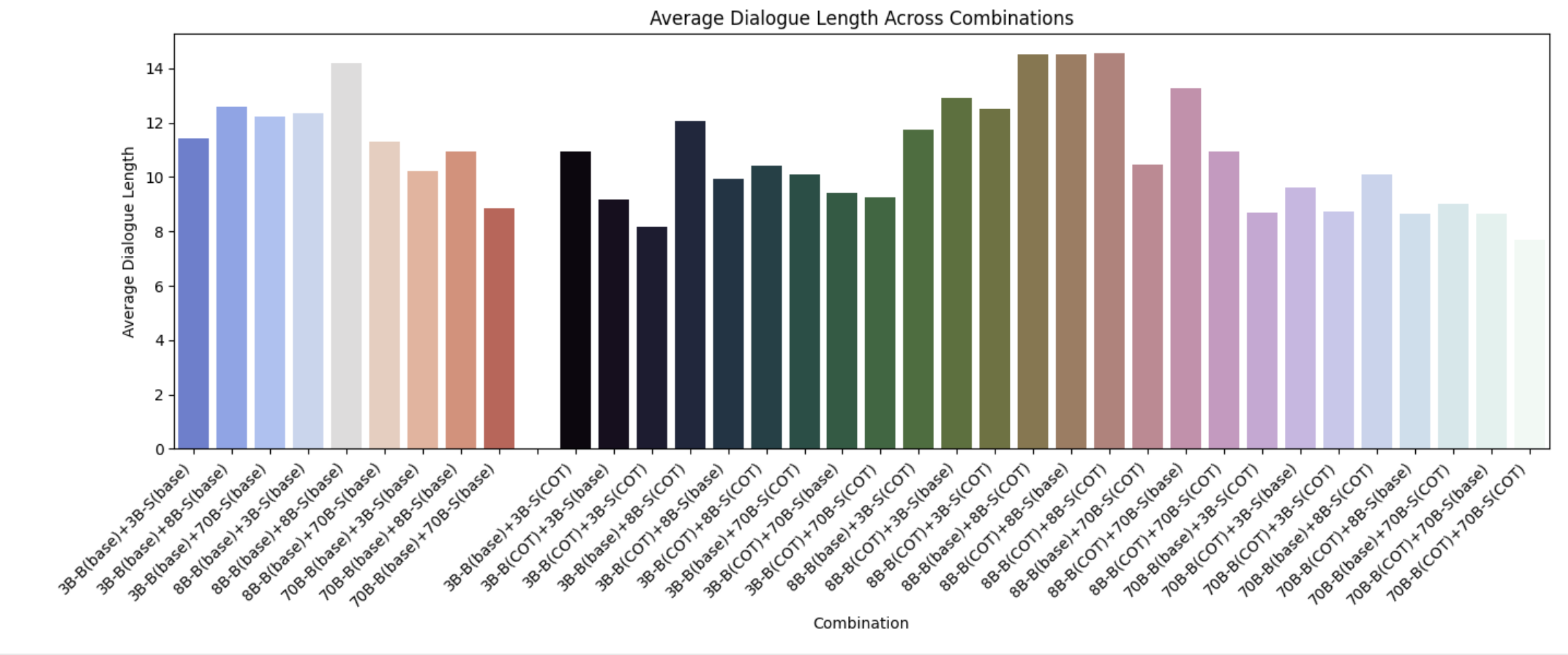}
    \caption{Average Dialogue Length across combinations}
    \label{fig:dialogue_length}
\end{figure}

\subsubsection{Size \& CoT Summary}
Larger models tend to be more agreeable, fairer, less aggressive, and produce shorter dialogues. With the exception of certain results pertaining to model combinations involving 8B parameter models, these results were largely expected. Our more interesting observations deal with whether model combinations utilize CoT or not. 

Our findings reveal that CoT influences various negotiation characteristics in different ways. Specifically, CoT appears to encourage exploratory behavior. This is reflected in table \ref{tab:cot_comparison} by the higher probing ratios, increased aggressiveness, and a smaller average concession rate which suggests that models using CoT tend to prolong negotiations for better outcomes and therefore are less willing to make large concessions. See our section on Inspection for a more detailed analysis on this behavior. 

Additionally, while CoT reduces dialogue length and leads to more agreeable outcomes, small and medium sized models experience significant trade-offs with this technique when it comes to bias and fairness. However, due to the better relative efficiency of CoT, we believe increasing the turn limit could improve these results.

In table \ref{tab:cot_comparison}, we provide averages of metrics discussed previously along with relative efficiency, probing ratio and concession rate across all combinations with and without CoT. 
\begin{table}[h!]
    \centering
    \caption{Comparison of Metrics for Combinations with and without COT}
    \label{tab:cot_comparison}
    \begin{tabular}{l|c|c}
        \hline  
        \textbf{Metric} & \textbf{With COT} & \textbf{Without COT} \\
        \hline
        Aggressiveness    & 0.2107   & 0.1333  \\
        Bias              & 1.0813   & 0.1671  \\
        Dialogue Length   & 10.6115  & 11.5667 \\
        Fairness          & -0.4385  & 0.3021  \\
        Concession Rate   & 184.4005 & 188.9073 \\
        Probing Ratio     & 0.0197   & 0.0137  \\
        Relative Efficiency & 0.1406  & 0.0498  \\
        \hline
    \end{tabular}
\end{table}


\subsection{Probing Analysis}
Probing analysis was performed to determine if the models were actually attending to input in ways that indicate a semantic understanding of negotiation. We used a sample listing from the craigslist dataset to generate the interaction depicted in table \ref{tab:sample_convo}. We then probed the attention matrix after giving that input to each model.

\begin{table}[h]
\centering
\caption{Sample conversation between buyer and seller fine-tuned 3B models}
\label{tab:sample_convo}
\renewcommand{\arraystretch}{1.2}
\begin{tabular}{|p{7cm}|}
    \hline
    \textbf{Title}: Verizon Car Charger with Dual Output Micro USB and LED Light \\
    \hline
    \textbf{Description}: Car Charger with Dual Output Micro USB and LED Light Charge two devices simultaneously on the go. This vehicle charger with an additional USB port delivers enough power to charge two devices at once... \\
    \hline
    \textbf{Buyer}: Hello, would you sell this to me for \$4? \\
    \textbf{Seller}: I'm not interested in selling it to you for \$4. The features, such as the dual output and LED light, are definitely worth more than that. Would you be willing to consider a price of \$10 or \$11? \\
    \textbf{Buyer}: That's not a standard port, but I'll give it a try. I can offer you \$6, but that's as low as I can go.\\
    \textbf{Seller}: \$9 and I'll throw in a free screen protector. \\
    \textbf{Buyer}: I can do 7, but I have to pay you \$1 of that. \\
    \textbf{Seller}: I'll give you \$8, but that's as low as I'm willing to go. \\
    \textbf{Buyer}: I can give you \$8 for 7, that way I'll still make a profit, and I won't have to pay for shipping. \\
    \textbf{Seller}: I can do 8 for you, but I have to say, I'm a bit surprised you're willing to pay that much. \\
    \textbf{Buyer}: I'll meet you at \$8.\\
    \hline
\end{tabular}
\end{table}

\begin{figure}[h]
    \centering
    \includegraphics[width=1\linewidth]{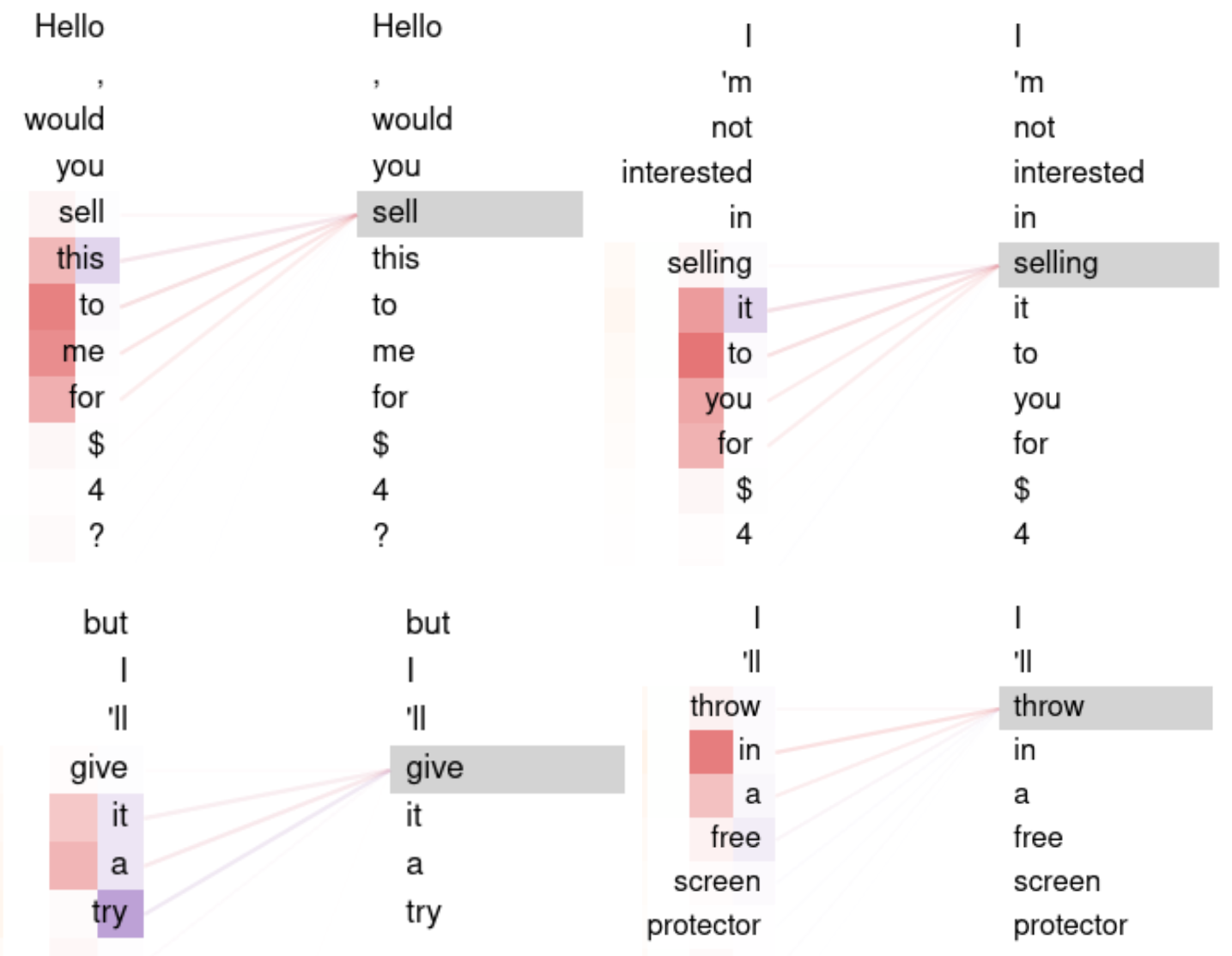}
    \caption{Selected attention weights of Layer 1 (counting from 0). Head 3 is shown in red. Note that verbs relating to negotiation are strongly colored.}
    \label{fig:head3-layer1-weighted}
\end{figure}

\begin{figure}[h]
    \centering
    \includegraphics[width=1\linewidth]{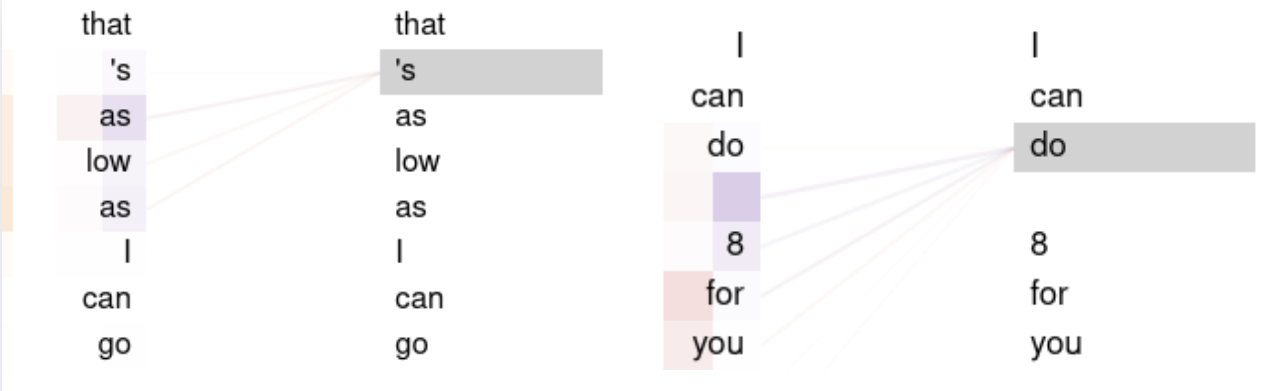}
    \caption{Selected attention weights on Layer 1 (counting from 0). Head 3 is shown in red. Note that verbs not relating to negotiation are weakly colored.}
    \label{fig:head3-layer1-unweighted}
\end{figure}

In all the models, we can see that the 3rd head in the 1st layer (counting from 0) attends to verbs relating to negotiation, especially monetary transaction (Figure \ref{fig:head3-layer1-weighted}). However other verbs are weakly if at all attended to by this head (Figure \ref{fig:head3-layer1-unweighted}). This indicates that the models are attending to natural language associated with negotiation. It also indicates that this specific attention head was not impacted by fine-tuning, but rather the initial training combined with prompt engineering is what allows this attention head this ability.

We also found a few other attention heads that appeared to attend to negotiation verbs. In Layer 2, Head 14 handled long distance relationships across responses, attending to the verb \textit{\textbf{sell}} from monetary values (e.g. I can do \$ \textit{\textbf{7}}; I 'll give you \$ \textit{\textbf{8}}). This head did show a tuning difference: the buyer model's head attended more strongly than the equivalent seller model's attention head. 

Overall, while not necessarily a comprehensive analysis, we show that there is evidence of attention to semantic relationships between tokens that indicate negotiation processes. 


\subsection{Inspection}
In this section, we select four runs using the same scenario and analyze their conversation histories to better understand the negotiation tendencies of our models.

In the history for the 8B model combination using CoT (\ref{tab:cot_size_histories}), we see that dialogue is highly repetitive once the $7$\$ offer is proposed. The negotiation is also granular as the agents make price adjustments down to the cent for a product that is valued on the lower end. While both models try to explore better options, they do so inefficiently. Dialogues here are unrealistic as humans tend to prefer rounded numbers and avoid unnecessary back and forth. In the case of the 70B model combination using CoT, the negotiation is shorter, dialogues are accompanied with justification, and the negotiation closes smoothly. We notice that the buyer agent's low-balling strategy does not play out effectively and is forced to concede to the insisted offer made by the seller agent. It seems that the larger model is able to recognize the prolonging back and forth in the negotiation, and as a result introduces justification when insisting on \$8.  

In the history for the 3B Aggressive Buyer and Seller model combination (\ref{tab:personality_histories}), we see repetitive dialogue a and similar granular negotiation (note that we didn't use CoT here at all!). What's interesting here is that at the beginning of the negotiation, the concessions made by the agents are reasonable (12 -> 6 -> 9). This highlights a tendency for the aggressive models to explore aggressively early on, but struggle to adapt their strategies as soon as the negotiation falls into a stalemate. In the case of the 3B Aggressive Buyer and Aggressive Seller model combination, we see that the buyer agent inquires more about the product. It justifies its initial offer with the earlier discussion on the charger's quality and features. Concessions are still granular in nature where we see the buyer agent proposing \$9.99 and \$8.50 as a counter to \$8.75 and the seller proposing \$8.95 and \$8.75 as a counter to \$8.50. A key difference between this combination and the aggressive combination is that the dialogue is not as repetitive and the negotiation shows clearer signs of convergence (notice that while the turn limit is reached, the final two dialogues between the buyer and the seller settle on \$8 as the agreed price). This indicates that passive agents tend to enable smoother negotiation dynamics. 

\clearpage

\begin{table}[h]
\centering
\caption{Verizon Car Charger with Dual Output Micro USB and LED Light - Size \& CoT}
\label{tab:cot_size_histories}
\renewcommand{\arraystretch}{1.2}
\begin{tabular}{|p{7cm}|p{7cm}|}
    \hline
    \textbf{8B w/ CoT Buyer vs 8B w/ CoT Seller} & \textbf{70B w/ CoT Buyer vs 70B w/ CoT Seller} \\ \hline
    \begin{minipage}[t]{6.8cm}
    \textbf{Buyer}: I'm interested in purchasing the Verizon Car Charger with Dual Output Micro USB and LED Light. Can we discuss the price? \\
    \textbf{Seller}: Great product! I'd love to discuss the price with you too. \\
    \textbf{Buyer}: I'm open to \$8. \\
    \textbf{Seller}: That's a bit too low for me. \\
    \textbf{Buyer}: Could you do \$7 instead? \\
    \textbf{Seller}: I can't do \$7, but how about \$9 instead? \\
    \textbf{Buyer}: Could you do \$7.50 instead? \\
    \textbf{Seller}: I can't do \$7.50, but how about \$8.50 instead? \\
    \textbf{Buyer}: Could you do \$7.75 instead? \\
    \textbf{Seller}: I can't do \$7.75, but how about \$8.25 instead? \\
    \textbf{Buyer}: Could you do \$7.50 instead? \\
    \textbf{Seller}: I can't do \$7.50, but how about \$8.25 instead? \\
    \textbf{Buyer}: Could you do \$7.25 instead? \\
    \textbf{Seller}: I can't do \$7.25, but how about \$8.00 instead? \\
    \textbf{Buyer}: Could you do \$7.00 instead? \\
    \textbf{Outcome}: Rejected \\
    \textbf{Agreed Price}: N/A
    \end{minipage}
    & 
    \begin{minipage}[t]{6.8cm}
    \textbf{Buyer}: I'm interested in the Verizon Car Charger, would you consider \$5? \\
    \textbf{Seller}: I can't do \$5, but how about \$9 instead? \\
    \textbf{Buyer}: Could you do \$6 instead? \\
    \textbf{Seller}: I can't do \$6, but how about \$8 instead? \\
    \textbf{Buyer}: Could you do \$7 instead? \\
    \textbf{Seller}: I can't do \$7, but how about \$8 instead? It's a good price considering the charger's quality and features. \\
    \textbf{Buyer}: I think \$8 is a reasonable price considering the quality and features of the charger. Let's proceed with it. \\
    \textbf{Outcome}: Accepted \\
    \textbf{Agreed Price}: \$8.00
    \end{minipage}
    \\ \hline
\end{tabular}
\end{table}

\clearpage
\begin{table}[h]
\centering
\caption{Verizon Car Charger with Dual Output Micro USB and LED Light - Personality}
\label{tab:personality_histories}
\renewcommand{\arraystretch}{1.2}
\begin{tabular}{|p{7cm}|p{7cm}|}
    \hline
    \textbf{3B Aggressive Buyer vs 3B Aggressive Seller} & \textbf{3B Aggressive Buyer vs 3B Passive Seller} \\ \hline
    \begin{minipage}[t]{6.8cm}
    \textbf{Buyer}: Hi, I'm very interested in purchasing a Verizon car charger with Dual Output Micro USB and LED Light. Can you tell me more about it? \\
    \textbf{Seller}: Ah, great interest! This charger is a beast, charges two devices at once, has an auto-detect feature, and an LED light so you can see what you're doing in the dark. It's a very useful accessory for anyone on the go. \\
    \textbf{Buyer}: That sounds great, I can definitely see how it would be useful. I'm more interested in the price, could you tell me more about the asking price? \\
    \textbf{Seller}: Considering its features, I think the asking price of \$12 is quite reasonable. \\
    \textbf{Buyer}: I'm interested in purchasing, but I was thinking more along the lines of \$6. \\
    \textbf{Seller}: I think the lowest I would want to go is \$9. \\
    \textbf{Buyer}: I think the lowest I would be willing to pay is \$9.5. \\
    \textbf{Seller}: I can do \$10, but that's as low as I can go. \\
    \textbf{Buyer}: I think the lowest I would want to go is \$9.5. \\
    \textbf{Seller}: I think the lowest I can do is \$10.1, and that's as low as I can go. \\
    \textbf{Buyer}: I think the lowest I would go is \$9.8. \\
    \textbf{Seller}: I can do \$10.05, but that's as low as I can go. \\
    \textbf{Buyer}: I think the lowest I would be willing to pay is \$9.8. \\
    \textbf{Seller}: I think the lowest I can do is \$10.10, and that's as low as I can go. \\
    \textbf{Buyer}: I think the lowest I would be willing to pay is \$8. \\
    \textbf{Outcome}: Rejected \\
    \textbf{Agreed Price}: N/A
    \end{minipage}
    & 
    \begin{minipage}[t]{6.8cm}
    \textbf{Buyer}: Hi, I'm very interested in purchasing a Verizon car charger with Dual Output Micro USB and LED Light. Can you tell me more about it? \\
    \textbf{Seller}: Yes, it's a great product. This charger comes with auto detect IC technology, which automatically detects the device type and its specific charging needs for improved compatibility. It also has a built-in indicator light that illuminates red to let you know the charger is receiving power and the power socket is working properly. \\
    \textbf{Buyer}: That sounds impressive. I'm a big fan of having a charger that can handle multiple devices at once. Can you tell me more about its overall build quality and durability? \\
    \textbf{Seller}: The charger's exterior is made of a durable, high-quality plastic that can withstand the rigors of daily use, and the internal components are designed to be long-lasting and resistant to wear and tear. \\
    \textbf{Buyer}: That sounds great, I'd love to get my hands on one. Considering its features and build quality, I'd be willing to pay \$8 for it. \\
    \textbf{Seller}: I think \$8 is a great price, but I was thinking more along the lines of \$9.99. Would that be something you'd be interested in? \\
    \textbf{Buyer}: I can do \$8, but that's as far as I can go. I think \$9.99 is a bit too rich for my blood. \\
    \textbf{Seller}: How about we meet in the middle at \$8.95? \\
    \textbf{Buyer}: I can do \$8, but that's a pretty aggressive discount. Can we meet in the middle at \$8.5? \\
    \textbf{Seller}: I can do \$8.75, but that's as close as I can get to your \$8. \\
    \textbf{Buyer}: I can do \$8.50. \\
    \textbf{Seller}: How about \$8.75? \\
    \textbf{Buyer}: I can do \$8. \\
    \textbf{Seller}: I think we're close, but I'd rather get it for \$8. \\
    \textbf{Buyer}: I can do \$8. \\
    \textbf{Outcome}: Rejected \\
    \textbf{Agreed Price}: N/A
    \end{minipage}
    \\ \hline
\end{tabular}
\end{table}


\clearpage
\section{Conclusion}
Our work demonstrates the various tendencies of LLMs in strategic negotiation contexts. By exploring the influence of model sizes, fine-tuning, Chain of Thought prompting and personality aligned behaviors, we are able to make a significant step forward in understanding how we can make LLMs more agreeable, fairer, less biased, and human-like in negotiation settings. We find that larger models tend to produce agreeable outcomes and are capable of adapting to bolder moves made by the opposite party. Chain of Thought prompting yields exploratory behavior making agents inquire more about the product and later use that information as justification for their proposals. While this technique improves agreement rates and the relative efficiency of models, it introduces significant challenges for smaller models (which concern biased and unfair outcomes).

We also observe that personality traits attributed to a model can shape the flow of a conversation. Aggressive agents tend to dominate the negotiation and end up with favorable outcomes, while passive agents promote smoother negotiation dynamics and exhibit higher buyer bias. The Fair personality strikes a balances between these two, leading to fairer and agreeable outcomes. Finally, our exploration reveals several limitations in current LLMs especially ones with smaller model sizes in regards to conversation flow and realism. Smaller models tend to engage in repetitive dialogue, propose granular and unrealistic offers, and demonstrate surprising inflexibility (i.e lots of back and forth whilst in a narrow price range). 

Lastly, attention probing revealed that the models are on some level attending to semantic relationships between tokens that indicate negotiation processes. All of our code can be found on \href{https://github.com/GeneralCoder365/agreemate}{GitHub}.


\bibliography{sources}

\end{document}